\documentclass[11pt]{article}
\usepackage{acl}
\usepackage{times}
\usepackage{latexsym}
\usepackage{booktabs}
\usepackage{hyperref}
\usepackage{graphicx}
\usepackage{xcolor}
\usepackage{tikz}
\usepackage{placeins}
\usepackage{float}
\usepackage{fancyhdr}
\usetikzlibrary{arrows.meta,backgrounds,calc,fit,positioning}
\usepackage[T1]{fontenc}
\usepackage[utf8]{inputenc}
\usepackage{microtype}

\fancypagestyle{firstpagefooter}{%
  \fancyhf{}%
  \fancyfoot[L]{\footnotesize Code and artifacts: \url{https://github.com/suraj-ranganath/PIPE-Cypher/}\\
  Correspondence: \href{mailto:suranganath@ucsd.edu}{\texttt{suranganath@ucsd.edu}}}%
}

\title{PIPE-Cypher: Automatic Enterprise Benchmark Generation for Text-to-Cypher Systems}
\author{
Suraj Ranganath\\
Hal{\i}c{\i}o\u{g}lu School of Data Science and Computing\\
University of California, San Diego
\And
Anish Raghavendra\\
Independent Researcher
}

\begin{document}
\maketitle
\thispagestyle{firstpagefooter}

\begin{abstract}
Enterprise property graphs vary widely in schema structure, internal terminology, domain assumptions, governance constraints, and user interaction patterns. A deployment-relevant Text2Cypher benchmark therefore reflects the questions users and agents actually ask of that graph. Creating such a benchmark is difficult because schemas and values are unique, and graph structure changes over time. Each NL-query pair must also be executable, use real graph entities, preserve diversity, and remain balanced across query types and difficulty levels. We present PIPE-Cypher, a local benchmark-generation pipeline that turns a live property graph and optional seed queries from customer questions, analyst logs, or agent tool calls into balanced NL-to-Cypher benchmarks. PIPE-Cypher combines schema profiling, reverse-query grounding, constrained generation, deterministic Cypher governance, execution validation, redaction, diversity controls, and a calibrated local LLM judge. Using local Qwen3.5-9B generation and judging, PIPE-Cypher exports 3,000 accepted FinBench/SNB examples, completes three audited ablation suites, calibrates judge behavior with human labels, and evaluates 11 local downstream models. The resulting benchmark is deliberately discriminative: zero-shot transfer is weak, while a few-shot control shows that schema-specific example banks can help compatible model families. Together, PIPE-Cypher makes Text2Cypher benchmarking a repeatable process that evolves with the graph, its users, and its target workloads.
\end{abstract}

\section{Introduction}
Property graphs are attractive in enterprise settings because the facts of interest are often relational paths: account transfers, identity entitlements, access chains, customer interactions, supplier dependencies, and fraud rings. Cypher gives analysts a compact language for these patterns. As LLMs become natural-language interfaces for graph analytics, an organization cannot evaluate a Text2Cypher system only on public schemas; it needs to know whether the model handles its labels, relationship directions, values, governance rules, and recurring operational questions.

For evaluation, that privacy boundary matters as much as model accuracy. A static public dataset is useful for shared comparison, but it cannot contain a bank's account taxonomy, an identity team's permission graph, or the categorical values that make a query answerable. It also cannot change when the production schema changes. Industry teams therefore need something different: a repeatable way to turn a live property graph into a balanced, executable, privacy-aware NL-to-Cypher benchmark without sending sensitive schema or values to paid generation APIs.

At first glance, benchmark generation looks like a natural job for a general AI agent: inspect the schema, draft Cypher, run it, fix mistakes, and write questions. That can work for a small one-off study when a strong model has the right tools, prompts, and examples. It is a weaker recipe for enterprise refreshes. Graphs change as products ship, integrations are added, and analysts ask new questions; a benchmark factory has to run again at scale. Many deployments also prefer efficient local models, including quantized variants that run quickly on less hardware, because cost, latency, privacy, and availability matter. These models can often write useful individual queries, but they are less reliable at managing the full loop: grounding real values, preserving relationship directions, avoiding unsafe clauses, balancing categories and difficulty, rejecting ambiguous examples, and leaving evidence for review. We therefore make benchmark generation a constrained pipeline: the model handles language and Cypher generation, while deterministic graph checks make the process repeatable, scalable, and auditable.

PIPE-Cypher profiles a target graph, finds graph values that make candidate questions answerable, constrains a local generator, validates and repairs the generated Cypher, executes it, and then asks a local LLM judge to review only candidates that already have execution evidence. We treat Cypher correctness as something to check, not something to hope the prompt induces: relationship direction, read-only safety, exact literal use, categorical values, contextual return columns, and \texttt{RETURN DISTINCT} are enforced before examples are accepted.

\noindent\textbf{Contributions.} We make four contributions: (1) a local-model workflow for generating private NL-to-Cypher benchmarks from an organization's own graph; (2) outcome-aware reverse grounding and Cypher-specific validators for read-only safety, relationship direction, exact literals, categorical values, contextual returns, and conservative rewrites; (3) a scaled public-proxy evaluation over FinBench, SNB, and ICIJ with ablations, judge calibration, redaction audits, and an 11-model local transfer study; and (4) reproducibility artifacts for onboarding, value sampling, benchmark refresh, evidence packaging, and appendix-level audit.

\section{Related Work}
Recent Text2Cypher resources, including Mind the Query~\citep{mindthequery2025}, SyntheT2C~\citep{synthet2c2025}, Auto-Cypher~\citep{autocypher2025}, Text2Cypher~\citep{text2cypher2025}, CypherBench~\citep{cypherbench2025}, and the public Text2Cypher-2024 corpus~\citep{text2cypher2024dataset}, show that good Cypher data needs schema grounding, execution checks, verification, and complexity-aware evaluation. PIPE-RDF~\citep{ranganath2026piperd} makes a related benchmark-factory argument for RDF/SPARQL, using reverse querying, category-balanced generation, retrieval, deduplication, execution validation, and deployment metrics. For enterprise Cypher benchmark generation, PIPE-Cypher adds outcome-aware reverse grounding before natural-language realization, deterministic property-graph governance before export, local judge calibration after execution, explicit privacy and value policies, and provenance-rich refresh artifacts for organizations that need to benchmark their own graphs.

Mind the Query is especially relevant because it reports an Industry Track Text2Cypher dataset with schema, runtime, value, and human logical validation. PIPE-Cypher keeps that validation discipline but changes the object of study. We are not proposing one more static dataset or tuning corpus. We are proposing the process an organization would use to generate, refresh, audit, and redact a benchmark for its own graph, local model endpoint, and value policy.

Text-to-SQL benchmarks such as Spider 2.0~\citep{spider22024} and BIRD~\citep{bird2023} have pushed text-to-query evaluation toward realistic database tasks and execution-based scoring, while execution-guided decoding shows why query execution is useful semantic feedback rather than a cosmetic check~\citep{wang2018executionguidedtexttosql}. Recent residual-skill Text-to-SQL work further shows that optimizing complementary agent skills on residual failures can improve selected accuracy across SQL dialects~\citep{zhu2026residualskilloptimizationtexttosql}. Graph query generation adds different failure modes: relationship direction can invert the meaning of a query, node and relationship properties live in different namespaces, and path-shaped questions can be syntactically valid while semantically ungrounded. CIKM AutoQuery~\citep{autoquery2024} motivates treating workload generation as a separate object of study from downstream model quality. LDBC FinBench~\citep{ldbcfinbench} and SNB~\citep{ldbcsnb} provide public graph workloads with financial and social-network structure, while ICIJ Offshore Leaks gives an additional public finance/compliance onboarding check.

For benchmark quality, lexical diversity alone is not enough. A question bank can use varied wording and still overuse the same graph values or the same Cypher template. PIPE-Cypher therefore combines text-generation diagnostics such as Distinct-n~\citep{li2016diversity} and self-BLEU-style redundancy checks~\citep{zhu2018texygen} with text-to-query structure metrics: schema coverage, relationship/property coverage, query-signature diversity, structural feature rates, and normalized entropy over graph/category/difficulty cells. For subset construction, we use an MMR-style novelty objective~\citep{carbonell1998mmr} over Cypher signatures, template families, structural substructures, schema atoms, values, and question tokens.

\section{Method}
PIPE-Cypher has six stages: schema profiling, workload planning, reverse Cypher grounding, constrained generation and repair, deterministic validation and execution, and LLM-judge review. The central design choice is simple: accepted examples should prove that they are answerable and safe. Prompts can ask a model to respect relationship directions or exact literals, but accepted examples must pass schema checks, parser-style structure extraction, live execution, and judge review.

Schema profiling records labels, relationship types, properties, observed directions, and bounded low-cardinality categorical values. Workload planning targets eight categories that appear in operational graph analytics: simple retrieval, complex retrieval, simple aggregation, complex aggregation, boolean existence, negation/difference, path/temporal transaction, and ranking/top-k. Reverse grounding then runs read-only Cypher to find slot values that actually produce rows. This step avoids a common synthetic-data failure: a plausible-looking question that has no answer in the graph.

The deterministic layer checks read-only safety, syntax shape, labels and properties from the schema, explicit relationship types, observed relationship directions, schema-provided categorical values, and non-empty execution where required. Live execution uses read-only credentials and read-access sessions; token-level write rejection is only the first safety check. A lightweight Cypher analyzer extracts return aliases, variables, labels, relationship patterns, risky constructs, and rewrite skip reasons. This makes normalization auditable instead of a silent string edit. The direction gate reads both outgoing and incoming arrow syntax before schema checking, and rejects undirected relationship patterns when direction is required.

\begin{figure*}[t]
\centering
\resizebox{\textwidth}{!}{\definecolor{pipeBlue}{HTML}{2F5F98}
\definecolor{pipeTeal}{HTML}{00856F}
\definecolor{pipeGold}{HTML}{B87A00}
\definecolor{pipeRed}{HTML}{B5524A}
\definecolor{pipePurple}{HTML}{6B5CA5}
\definecolor{pipeInk}{HTML}{243040}
\definecolor{pipeSlate}{HTML}{5A6778}
\definecolor{pipeLine}{HTML}{AAB4C0}
\definecolor{pipePaleBlue}{HTML}{EAF1F8}
\definecolor{pipePaleTeal}{HTML}{E8F5F1}
\definecolor{pipePaleGold}{HTML}{FBF2DA}
\definecolor{pipePaleRed}{HTML}{F8EAE8}
\definecolor{pipePalePurple}{HTML}{F0EDF8}
\definecolor{pipePaleGray}{HTML}{F3F5F7}
\providecommand{\pipefigitem}[1]{\raisebox{0.18ex}{\tiny$\bullet$}\hspace{0.24em}#1\\}
\providecommand{\pipefiglastitem}[1]{\raisebox{0.18ex}{\tiny$\bullet$}\hspace{0.24em}#1}

\begin{tikzpicture}[
  font=\sffamily,
  >=Latex,
  stage/.style={
    rounded corners=2pt,
    draw=pipeInk,
    line width=0.55pt,
    minimum height=0.92cm,
    text width=2.18cm,
    align=center,
    inner sep=4pt,
    font=\sffamily\scriptsize\bfseries
  },
  detail/.style={
    rounded corners=1.5pt,
    draw=pipeLine,
    line width=0.35pt,
    fill=white,
    text width=2.18cm,
    align=left,
    inner sep=3pt,
    font=\sffamily\tiny
  },
  sidebox/.style={
    rounded corners=2pt,
    draw=pipeLine,
    line width=0.45pt,
    fill=pipePaleGray,
    text width=2.15cm,
    align=left,
    inner sep=4pt,
    font=\sffamily\tiny
  },
  loopbox/.style={
    rounded corners=2pt,
    draw=pipeSlate,
    line width=0.45pt,
    fill=white,
    text width=3.05cm,
    align=left,
    inner sep=4pt,
    font=\sffamily\tiny
  },
  contextbox/.style={
    rounded corners=2pt,
    draw=pipeSlate,
    line width=0.45pt,
    fill=white,
    text width=2.33cm,
    align=left,
    inner sep=4pt,
    font=\sffamily\tiny
  },
  artifact/.style={
    rounded corners=2pt,
    draw=pipeTeal,
    line width=0.55pt,
    fill=pipePaleTeal,
    text width=2.45cm,
    align=left,
    inner sep=4pt,
    font=\sffamily\tiny
  },
  arrow/.style={->, line width=0.75pt, draw=pipeInk},
  softarrow/.style={->, line width=0.55pt, draw=pipeSlate},
  rejectarrow/.style={->, line width=0.6pt, draw=pipeRed, dashed},
  refresharrow/.style={->, line width=0.65pt, draw=pipeTeal, dashed}
]

\node[stage, fill=pipePaleBlue] (schema) at (0,0)
  {1. Schema and\\value profiling};
\node[stage, fill=pipePaleGold, right=0.55cm of schema] (plan)
  {2. Workload\\planning};
\node[stage, fill=pipePaleTeal, right=0.55cm of plan] (ground)
  {3. Reverse\\grounding};
\node[stage, fill=pipePalePurple, right=0.55cm of ground] (generate)
  {4. Constrained\\generation};
\node[stage, fill=pipePaleRed, right=0.55cm of generate] (govern)
  {5. Cypher\\governance};
\node[stage, fill=pipePaleBlue, right=0.55cm of govern] (execute)
  {6. Execution\\validation};
\node[stage, fill=pipePaleTeal, right=0.55cm of execute] (judge)
  {7. Judge, audit,\\and export};

\draw[arrow] (schema) -- (plan);
\draw[arrow] (plan) -- (ground);
\draw[arrow] (ground) -- (generate);
\draw[arrow] (generate) -- (govern);
\draw[arrow] (govern) -- (execute);
\draw[arrow] (execute) -- (judge);

\node[detail, below=0.18cm of schema] (schemaDetails) {
  \textbf{Artifacts}\\
  \pipefigitem{labels / rel types}
  \pipefigitem{properties}
  \pipefigitem{directions}
  \pipefigitem{categorical values}
  \pipefiglastitem{schema fingerprint}
};
\node[detail, below=0.18cm of plan] (planDetails) {
  \textbf{Controls}\\
  \pipefigitem{graph targets}
  \pipefigitem{category targets}
  \pipefigitem{difficulty buckets}
  \pipefigitem{diversity caps}
  \pipefiglastitem{attempt budget}
};
\node[detail, below=0.18cm of ground] (groundDetails) {
  \textbf{Grounding}\\
  \pipefigitem{reverse bindings}
  \pipefigitem{entity/value slots}
  \pipefigitem{exact literal policy}
  \pipefiglastitem{placeholder examples}
};
\node[detail, below=0.18cm of generate] (genDetails) {
  \textbf{Local model}\\
  \pipefigitem{Qwen3.5-9B endpoint}
  \pipefigitem{schema-only prompts}
  \pipefigitem{Cypher constraints}
  \pipefiglastitem{local generation only}
};
\node[detail, below=0.18cm of govern] (govDetails) {
  \textbf{Deterministic gates}\\
  \pipefigitem{read-only safety}
  \pipefigitem{parser/AST features}
  \pipefigitem{schema vocabulary}
  \pipefiglastitem{direction/value checks}
};
\node[detail, below=0.18cm of execute] (execDetails) {
  \textbf{Live Cypher run}\\
  \pipefigitem{syntax/runtime checks}
  \pipefigitem{result samples}
  \pipefigitem{non-empty gate}
  \pipefiglastitem{empty-result diagnosis}
};
\node[detail, below=0.18cm of judge] (judgeDetails) {
  \textbf{Quality review}\\
  \pipefigitem{relevant schema + result sample}
  \pipefigitem{local LLM-judge JSON}
  \pipefigitem{ambiguity/alignment}
  \pipefigitem{human-audit}
  \pipefiglastitem{calibration}
};

\node[sidebox, above=0.78cm of schema] (inputs) {
  \textbf{Enterprise inputs}\\
  \pipefigitem{private property graph}
  \pipefigitem{read-only Cypher endpoint}
  \pipefigitem{domain terminology}
  \pipefiglastitem{local model endpoint}
};
\node[sidebox, above=0.78cm of ground] (policy) {
  \textbf{Privacy policy}\\
  \pipefigitem{value sampling bounds}
  \pipefigitem{sensitive-property omit list}
  \pipefigitem{redaction rules}
  \pipefiglastitem{export visibility level}
};
\draw[softarrow] (inputs.south) -- (schema.north);
\draw[softarrow] (policy.south west) -- (schema.north east);
\draw[softarrow] (policy.south) -- (ground.north);

\node[contextbox, above=0.78cm of generate] (retrieval) {
  \textbf{Retrieval memory}\\
  accepted examples with tenant values replaced by typed placeholders; nearest examples feed few-shot prompts without leaking raw values
};
\draw[softarrow] (ground.north east) to[out=35,in=180] (retrieval.west);
\draw[softarrow] (retrieval.south) -- (generate.north);

\node[contextbox, above=1.26cm of govern] (rewrite) {
  \textbf{Repair and rewrite}\\
  \pipefigitem{add \texttt{RETURN DISTINCT}}
  \pipefigitem{normalize projections}
  \pipefiglastitem{reject risky \texttt{CALL}, \texttt{UNION}, writes, reserved variables, or parser-risky cases}
};
\draw[softarrow] (govern.north) -- (rewrite.south);
\draw[softarrow] (rewrite.east) to[out=0,in=135] (execute.north west);

\node[artifact, right=0.62cm of judge] (export) {
  \textbf{Benchmark package}\\
  \pipefigitem{NL-Cypher JSONL}
  \pipefigitem{train/dev/test splits}
  \pipefigitem{result samples}
  \pipefigitem{difficulty features}
  \pipefigitem{benchmark card}
  \pipefigitem{refresh manifest}
  \pipefiglastitem{redacted review export}
};
\draw[arrow] (judge) -- (export);

\node[loopbox, below=2.58cm of govern] (ledger) {
  \textbf{Rejected-candidate ledger}\\
  \pipefigitem{validation issues}
  \pipefigitem{judge failures}
  \pipefigitem{duplicate/diversity blocks}
  \pipefigitem{repair decisions}
  \pipefigitem{empty-result classes}
  \pipefiglastitem{prompt profile / run ID}
};
\node[loopbox, below=2.58cm of plan] (topup) {
  \textbf{Top-up and refresh controller}\\
  \pipefigitem{rebalances missing cells}
  \pipefigitem{graph/category/difficulty}
  \pipefigitem{estimates capacity}
  \pipefigitem{starts scaled runs}
  \pipefiglastitem{from logged gaps}
};
\draw[rejectarrow] (execDetails.south) to[out=-90,in=70] (ledger.north east);
\draw[softarrow] (ledger.west) -- (topup.east);
\coordinate (topupReturnX) at ($(schemaDetails.east)!0.5!(planDetails.west)$);
\draw[softarrow]
  (topup.north west) -- (topupReturnX |- topup.north west)
  -- (topupReturnX |- plan.south west)
  -- (plan.south west);

\begin{scope}[on background layer]
  \node[
    rounded corners=3pt,
    draw=pipeLine,
    fill=pipePaleGray,
    fit=(schema) (plan) (ground) (generate) (govern) (execute) (judge)
      (schemaDetails) (planDetails) (groundDetails) (genDetails) (govDetails) (execDetails) (judgeDetails),
    inner xsep=6pt,
    inner ysep=7pt
  ] (mainfit) {};
\end{scope}

\end{tikzpicture}}
\caption{PIPE-Cypher benchmark generation pipeline. The key industry additions beyond static Text2Cypher dataset construction are privacy/value policies, reverse grounding, Cypher governance, execution diagnostics, local judge calibration, and benchmark export/refresh.}
\label{fig:pipeline_overview}
\end{figure*}
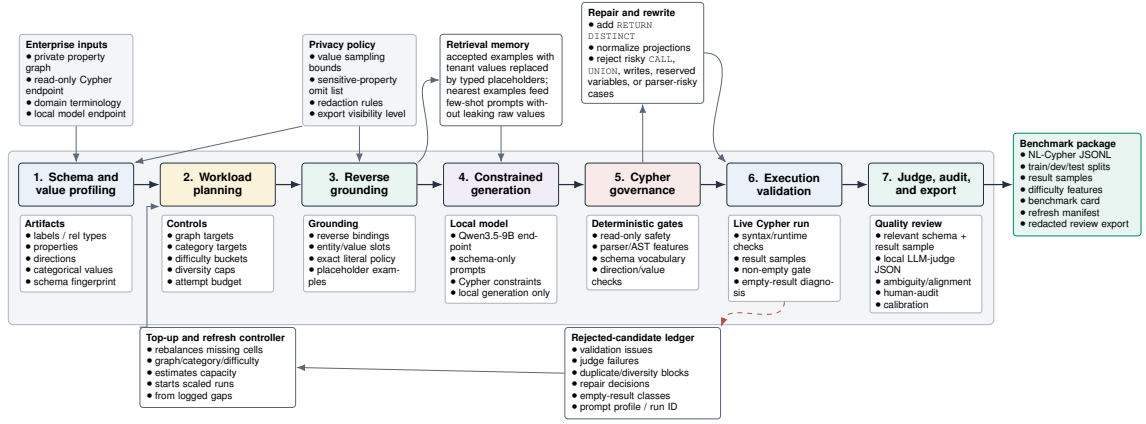

\begin{table*}[t]
\centering
\small
\begin{tabular}{p{0.22\textwidth}p{0.36\textwidth}p{0.34\textwidth}}
\toprule
Gate & Evidence checked & Failure mode caught \\
\midrule
Read-only safety & blocked Cypher write/admin tokens & destructive or operational queries \\
Schema validity & labels, relationship types, properties, categorical values & hallucinated graph vocabulary or values \\
Direction validity & observed relationship directions & reversed or invalid traversals \\
Cypher analysis and normalization & structure extraction, rewrite safety, \texttt{RETURN DISTINCT} & unsafe rewrites or duplicate/noisy rows \\
Question constraints & quoted values use exact literals & fuzzy matching of exact user values \\
Execution & query runs and returns rows & syntactic validity without answerability \\
LLM judge & ambiguity, semantic alignment, schema use & executable but semantically weak examples \\
\bottomrule
\end{tabular}
\caption{PIPE-Cypher candidate acceptance gates.}
\label{tab:gates}
\end{table*}

\section{Implementation}
PIPE-Cypher is a Python package built around a read-only graph client, a schema/value profiler, local model endpoints, Cypher governance, and export/audit tools. Before any model call, the profiler records the schema, relationship directions, and value samples the run may use. Generation, validation, judging, and export use that profile and execution traces rather than backend-specific objects. We use Neo4j for experiments, but only the graph client is backend-specific; the method targets Cypher over property graphs. Benchmark generation and judge review use a local Qwen3.5-9B endpoint~\citep{qwen2026qwen35hf} behind a vLLM/OpenAI-compatible interface. Downstream evaluation uses 11 completed locally served checkpoints from general instruction, code-tuned, Cypher-tuned, and Text2Cypher-tuned families. We keep these roles separate: one endpoint builds the benchmark, and the downstream study measures how other local models behave on the exported examples. All generation and evaluation stay inside the organization's compute boundary without paid generation APIs.

The graph profile makes graph-specific assumptions explicit. FinBench uses the public datagen snapshot export with typed node and relationship properties. SNB uses the official Neo4j/Cypher headers and read-query files. Live inspection also records bounded low-cardinality strings as categorical constraints. During FinBench import, we create rather than merge transaction relationships so repeated account-to-account events remain visible to path and aggregation queries. These choices determine answerability: a property may belong to a relationship rather than a node, a value may be unsafe to sample, and a relationship may only make sense in one direction. A company-owned onboarding run creates the same profile before the first LLM call.

For generation, PIPE-Cypher starts from workload templates whose slots are filled by reverse-binding queries. It can also use a mixed mode in which the LLM proposes additional templates after seeing proven workload seeds. Every run records both accepted and rejected candidates. Retrieved few-shot examples replace graph-specific values with typed placeholders, so the model sees the query structure without repeatedly seeing the same tenant values. A lightweight value grounder adds typed annotations for categorical values and reverse-bound entities, including punctuation variants, possessives, plurals, synonyms, name partials, and small typos.

Inspired by Mind the Query's prompt-setting analysis, PIPE-Cypher exposes prompt profiles for schema-only, instructions-only, examples-only, examples-plus-instructions, and full governed generation. We report a profile only when it passes the same target-size and evidence checks as the main run.

For LLM-judge review, we do not send the entire schema when a query touches only a small part of it. The judge prompt includes the labels, relationship types, and properties mentioned by the candidate query, while deterministic validators still check against the full schema. This keeps local 9B prompts manageable on larger graphs without weakening schema validation.

The Cypher layer uses constraints drawn from production Text2Cypher work: schema-only prompting, exact matching, relationship direction discipline, \texttt{RETURN DISTINCT}, reserved variable rejection, categorical values, required contextual return columns, fuzzy value annotations, placeholderized retrieval examples, and parser-aware rewrite boundaries. PIPE-Cypher records parser-style structure features and skips rewrites for risky constructs such as \texttt{UNION}, \texttt{CALL}, \texttt{UNWIND}, \texttt{WHERE EXISTS}, multiple \texttt{WHERE} clauses, or reserved variables.

Before scaling a run, PIPE-Cypher checks that each target category has enough executable slot bindings; this is only a launch guard, and exported examples still pass validation, execution, diversity, and judge gates.

For enterprise onboarding, PIPE-Cypher includes a deployment template for a company's own graph: read-only credentials, local model endpoint, schema introspection, privacy policy, dry run, scaled run, audit, and redacted export. We validate this pattern on three public proxy graphs rather than a proprietary tenant deployment. Configurable value-sampling policies decide which low-cardinality graph values may enter prompts. Redacted exports replace quoted literals, entity values, and string-valued result samples with stable placeholders for broader internal review. To support schemas beyond the built-in LDBC profiles, PIPE-Cypher derives relationship-count, anti-join, and top-k templates from observed labels, relationship directions, and safe low-cardinality properties, then grounds slot values with outcome-aware reverse Cypher.

\section{Experiments}
\noindent\textbf{Research questions.} We evaluate four questions that matter for an industry benchmark generator: RQ1, can a local-model pipeline produce a balanced executable benchmark over live property graphs? RQ2, do Cypher-specific validation and grounding steps make generation reliable at scale? RQ3, does the resulting benchmark expose meaningful downstream Text2Cypher failures rather than merely checking syntax? RQ4, can the same workflow onboard a new public enterprise-style graph without hard-coding FinBench or SNB?

The generated benchmark contains 3,000 accepted examples: 2,000 from LDBC FinBench and 1,000 from LDBC SNB, balanced over eight workload categories. We report three completed FinBench/SNB ablation suites: target-50, corrected target-100, and a seed-17 target-50 repeat. Downstream Text2Cypher evaluation uses live execution accuracy and answer-set F1 as primary metrics; reference-based text metrics are supported for debugging only and reported in the appendix. We additionally report ICIJ Offshore Leaks as a third public finance/compliance onboarding proxy.

\section{Results}
\noindent\textbf{RQ1: executable benchmark generation.} The full live run produced 3,000 accepted examples from 4,925 candidates using local Qwen3.5-9B for generation and judging. Category-specific recovery top-ups filled the only under-target categories from the initial sequential run. Every exported example passed read-only, syntax, schema, execution, non-empty result, and judge gates.

\begin{table}[t]
\centering
\small
\resizebox{\columnwidth}{!}{%
\begin{tabular}{lrrrr}
\toprule
Graph & Candidates & Accepted & Acceptance & Categories at target \\
\midrule
FinBench & 3,405 & 2,000 & 0.587 & 8/8 \\
SNB & 1,520 & 1,000 & 0.658 & 8/8 \\
Total & 4,925 & 3,000 & 0.609 & 16/16 \\
\bottomrule
\end{tabular}
}
\caption{Full live generation with local Qwen3.5-9B. Candidate counts include the initial sequential run and category-specific recovery top-ups.}
\label{tab:full_generation_results}
\end{table}

The accepted records are exported with stable identifiers, train/dev/test splits, result samples, gate metadata, aggregate statistics, and a manifest hash; the appendix gives the full artifact distribution.

\noindent\textbf{Diversity and residual concentration.} We do not reduce diversity to one score. The full export has perfect category balance, near-perfect difficulty balance, 1,115 unique grounded entity values, and exact quotation of grounded values in 82.6\% of examples with entity bindings. Query-signature diversity remains low because seeded graph-grounded templates carry much of the generation load. Table~\ref{tab:diversity_main} shows that diversity is still governable after acceptance: at the same graph/category target, a selector using query signatures, template families, structural substructures, schema atoms, values, and question tokens improves structural coverage, adjusted Distinct-2, query-signature ratio, and property coverage. The result is a balanced executable benchmark, plus clear diagnostics for what should diversify during refresh.

\begin{table}[H]
\centering
\small
\begin{tabular}{lrr}
\toprule
Metric & Random & Governed \\
\midrule
PIPE-Diversity index & 0.557 & 0.575 \\
Unique query-signature ratio & 0.062 & 0.135 \\
Structural substructures & 97 & 134 \\
Adjusted Distinct-2 & 0.266 & 0.284 \\
Property coverage & 0.407 & 0.426 \\
\bottomrule
\end{tabular}
\caption{Diversity-governed selection at the same graph/category target. The selector improves structural, lexical, signature, and schema coverage after all quality gates have already passed; the appendix reports the full diversity audit, including residual template concentration.}
\label{tab:diversity_main}
\end{table}

\noindent\textbf{RQ2: governed generation.} Most full-run rejections come from duplicate/diversity controls or empty execution results. Only 2 of 4,925 candidates were schema-invalid after the Cypher checks. A rewrite audit found that all reported-run candidates were already identical to their normalized Cypher, so accepted examples do not depend on semantics-changing rewrites. The ablations should therefore be read as a reliability study of the execution-grounded core: in the target-100 suite, every non-unconstrained graph/setting cell reached all eight category targets, while unconstrained generation did not produce balanced executable coverage. Across the three evidence-ready suites, target-normalized coverage is 1.000 for every non-unconstrained cell.

\noindent\textbf{Judge calibration.} An 80-row post-hoc human audit sampled accepted and rejected candidates across both graphs and all categories. Agreement is 80.0\%, Cohen's $\kappa=0.60$, judge precision/specificity are 1.00, recall is 0.714, and no false accepts appear in the sample. The judge is conservative and protects accepted-example quality; human labels calibrate the gate but do not participate in generation.

\noindent\textbf{RQ3: downstream stress test.} We evaluate downstream Text2Cypher by giving each model the schema text and the question, then scoring the generated query by live execution on FinBench or SNB. This is where the benchmark becomes useful: many outputs parse, mention plausible schema, and still answer the wrong graph question. The local Qwen3.5-9B baseline, for example, reaches 0.963 parse validity and 0.916 schema validity but only 0.189 exact execution accuracy on the 296-example held-out split. In an 11-model completed local transfer suite, zero-shot execution accuracy ranges from 0.000 to 0.203 (mean 0.036). Table~\ref{tab:downstream_transfer_summary_main} gives the control comparison. The primary few-shot generalization result is the scored no-signature control, which excludes exact query-signature matches and near-duplicate questions and raises mean accuracy to 0.200. Ordered and random same-category example banks reach 0.269/0.267, but we treat them as operational upper bounds because they often share query signatures. This split is important for industry use: the benchmark can be a hard model-evaluation set and, separately, a private example bank for schema-specific retrieval or adaptation.

\begin{table}[H]
\centering
\footnotesize
\begin{tabular}{@{}lccc@{}}
\toprule
Condition & Sig. & Mean & Best \\
\midrule
Zero-shot & -- & 0.036 & 0.203 \\
No-signature & 0.000 & 0.200 & 0.828 \\
Ordered & 0.866 & 0.269 & 0.993 \\
Random & 0.854 & 0.267 & 0.986 \\
\bottomrule
\end{tabular}
\caption{Eleven-model local downstream transfer controls on the 296-example held-out split. Sig. is the fraction of selected demonstrations sharing the test query signature; the no-signature row is the leakage-aware example-bank result, while ordered and random same-category rows are operational upper bounds.}
\label{tab:downstream_transfer_summary_main}
\end{table}

\noindent\textbf{RQ4: third-graph onboarding.} ICIJ onboarding reaches 800 accepted examples from 983 candidates on a 2.0M-node, 3.3M-edge public finance/compliance graph, with 100 examples in every category. This is evidence from a third public graph, not proof of private-tenant coverage. It is still important because it exercises schema-derived relationship-count, anti-join, and top-k templates beyond the two LDBC workloads.

\section{Industry Use}
Enterprise graphs are usually specialized artifacts, not generic benchmark schemas. They encode a company's products, risk rules, permissions, data integrations, and analyst vocabulary. A Text2Cypher system is useful only if it works on the questions users actually ask of that graph. PIPE-Cypher treats seed queries as a practical bridge from deployment to evaluation: when available, they can come from historical customer questions, analyst query logs, or agent tool calls triggered by user requests. The pipeline then expands those seeds into a balanced benchmark that tests the same kinds of operations the organization expects its agent to perform.

This matters because the downstream stress test in Table~\ref{tab:downstream_transfer_summary_main} and Appendix Figure~\ref{fig:downstream_fewshot_controls} shows that general Text2Cypher models often do not transfer cleanly to a new graph with its own schema. The generated examples are therefore useful in two ways. First, they form a private held-out test set for measuring agent behavior under the organization's schema, values, and safety rules. Second, accepted examples can become a schema-specific question--query bank for retrieval-augmented prompting; the no-signature few-shot control improves mean local-model accuracy, while same-category banks give an operational upper bound (Appendix Tables~\ref{tab:downstream_fewshot_controls}--\ref{tab:fewshot_leakage_controls}). For more complex deployments, the same accepted pairs can also seed supervised adaptation, although we do not claim a tenant-specific fine-tuning result here.

PIPE-Cypher is meant to be rerun when an enterprise graph changes. Schemas change as teams add products, ingest new sources, or encode new business logic, and static Text2Cypher corpora become stale quickly. Each example records the schema snapshot, graph profile, model identifier, validation gates, execution sample, judge scores, difficulty features, and source run, so the benchmark can be refreshed and audited. A deployment needs read-only graph credentials, schema introspection, a bounded value policy, a local endpoint, a dry run, scaled generation, judge calibration, and redacted export. Local inference keeps generation inside the compute boundary; if an organization later permits remote inference, the same value-sampling and redaction policies provide a safer artifact boundary. We validate this pattern on FinBench, SNB, and ICIJ while keeping the private-tenant gap explicit.

\section{Conclusion}
PIPE-Cypher reframes Text2Cypher benchmarking as a private, repeatable enterprise workflow. The main lesson is that generation improves when Cypher constraints become executable checks. Reverse grounding makes questions answerable. Deterministic validation catches unsafe or schema-invalid queries. Execution exposes empty or brittle candidates. Diversity diagnostics reveal concentration. A calibrated local judge adds a conservative semantic filter. Together these pieces produce a benchmark that is balanced, auditable, refreshable, and able to reveal downstream model failures that syntax-only evaluation would hide.

\section*{Limitations}
Execution validity does not guarantee semantic correctness. The completed 80-row, single-human-annotator calibration suggests a conservative judge with no observed false accepts in the labeled sample, but the confidence interval is wider than the point estimate and larger multi-annotator audits may reveal additional failure modes. FinBench, SNB, and ICIJ are public enterprise-style proxies rather than a proprietary tenant graph, so we test the onboarding pattern but not every deployment constraint of a real organization. The full export is balanced by graph, category, and difficulty, but query-signature diagnostics still show template concentration from seeded, execution-grounded generation. PIPE-Cypher deliberately disallows or skips risky Cypher constructs such as writes, undirected relationships, \texttt{UNION}, \texttt{CALL}, \texttt{UNWIND}, and parser-risky rewrites. This is a safe benchmark-generation subset, not complete coverage of every production Cypher idiom. The downstream few-shot result should be read as graph-specific example-bank conditioning: the no-signature control is the leakage-aware result, while ordered and random same-category demonstrations are upper-bound conditions that often share query signatures. Tenant-specific fine-tuning remains an engineering path enabled by the artifact, not a completed deployment claim here. The redaction audit checks exact residuals for known value-bearing strings, but it is not a full PII classifier and does not make schema names confidential.

\section*{Ethics Statement}
PIPE-Cypher is designed for private benchmark generation under local-model deployment constraints. Organizations using it should restrict benchmark artifacts to authorized users, review sampled values for sensitive content, and document whether judge calibration relied on human annotators. For this study, one external human annotator labeled an 80-row post-hoc judge-calibration packet. The annotator was informed that the labels would be used for research calibration and paper reporting, and raw value-bearing annotation rows are not released. The protocol received an IRB exemption; identifying review-board details are omitted for double-blind submission. Human labels were never used as a generation gate.

\bibliography{references}

\clearpage
\appendix
\raggedbottom

\section{Prior-Work Positioning}

Table~\ref{tab:prior_mechanism_comparison} expands the related-work comparison from the main paper. PIPE-Cypher's contribution is the way answerability, governance, privacy, judging, and refresh are assembled into an organization-run benchmark factory.

\begin{table}[H]
\centering
\footnotesize
\begin{tabular}{@{}p{0.37\columnwidth}p{0.56\columnwidth}@{}}
\toprule
Prior mechanism & PIPE-Cypher delta \\
\midrule
Template/LLM synthesis in SyntheT2C~\citep{synthet2c2025} and Auto-Cypher~\citep{autocypher2025} & Outcome-aware reverse grounding binds slots through live Cypher before NL realization. \\
Execution validation in Auto-Cypher~\citep{autocypher2025} and Mind the Query~\citep{mindthequery2025} & Execution is one gate in a ledger with direction, read-only, value, non-empty, and judge evidence. \\
Schema/value checks in Mind the Query~\citep{mindthequery2025} & Checks are packaged for private refresh with configurable value sampling and redacted exports. \\
PIPE-RDF enterprise benchmark factory~\citep{ranganath2026piperd} & Adapts the factory framing from RDF/SPARQL to property graphs, where direction, relationship properties, and path semantics require Cypher-specific governance. \\
Human logical review in Mind the Query~\citep{mindthequery2025} & Human labels calibrate a local LLM judge; they are not a generation gate. \\
Public static datasets and tuning corpora~\citep{text2cypher2025,cypherbench2025,text2cypher2024dataset} & Organization-run benchmark factory with manifests, refresh, and provenance audits. \\
\bottomrule
\end{tabular}
\caption{Prior mechanism comparison. PIPE-Cypher combines outcome-aware grounding, deterministic Cypher checks, local judge calibration, privacy policy, refresh support, and audit logs so organizations can generate their own benchmarks rather than only consume static datasets.}
\label{tab:prior_mechanism_comparison}
\end{table}

\section{Experimental Setup and Exported Benchmark}

Unless noted otherwise, the extended results use the same live FinBench/SNB export as the main results. Tables in this section pin down the scale, graph mix, split, and validation totals so the downstream and diversity analyses are tied to a single benchmark artifact.

\subsection{Benchmark Artifact Summary}

Tables~\ref{tab:experiment_plan}, \ref{tab:benchmark_export}, and \ref{tab:full_artifact_distribution} give the run configuration, export manifest summary, and gate/distribution details for the 3,000-example benchmark.

\begin{table}[H]
\centering
\small
\resizebox{\columnwidth}{!}{%
\begin{tabular}{ll}
\toprule
Setting & Value \\
\midrule
Accepted examples & 3,000 \\
Primary graph & LDBC FinBench, 2,000 examples \\
Secondary graph & LDBC SNB, 1,000 examples \\
Categories & 8 balanced categories, 375 each \\
Generation/judge model & Local Qwen3.5-9B \\
Execution backend & Neo4j Community, two live databases \\
Judge audit packet & 80 sampled accepted/rejected pairs \\
\bottomrule
\end{tabular}
}
\caption{Full live experimental setup used for the generated benchmark artifact.}
\label{tab:experiment_plan}
\end{table}

\begin{table}[H]
\centering
\small
\resizebox{\columnwidth}{!}{%
\begin{tabular}{lrrrrrr}
\toprule
Export artifact & Examples & FinBench & SNB & Train & Dev & Test \\
\midrule
Live full benchmark & 3,000 & 2,000 & 1,000 & 2,408 & 296 & 296 \\
\bottomrule
\end{tabular}
}
\caption{Accepted live full benchmark package with stable IDs, gate metadata, result samples, statistics, and manifest hash \texttt{cf274344be2abe7e}.}
\label{tab:benchmark_export}
\end{table}

\begin{table}[H]
\centering
\small
\resizebox{\columnwidth}{!}{%
\begin{tabular}{lr}
\toprule
Artifact property & Value \\
\midrule
Categories & 8 balanced categories \\
FinBench/category & 250 \\
SNB/category & 125 \\
Difficulty split & 1,569 easy / 1,431 medium \\
Used labels / rel. types & 16 / 17 \\
Read/syntax/schema/exec/judge gates & 3,000/3,000 \\
\bottomrule
\end{tabular}
}
\caption{Distribution and gate summary for the exported full benchmark artifact.}
\label{tab:full_artifact_distribution}
\end{table}

\FloatBarrier

\section{Governed Generation Evidence}

The central reliability question is whether reverse grounding and Cypher validation can fill every planned graph/category cell at target scale.

\subsection{Target-100 Stress Baseline}

\begin{figure}[H]
\centering
\includegraphics[width=\columnwidth]{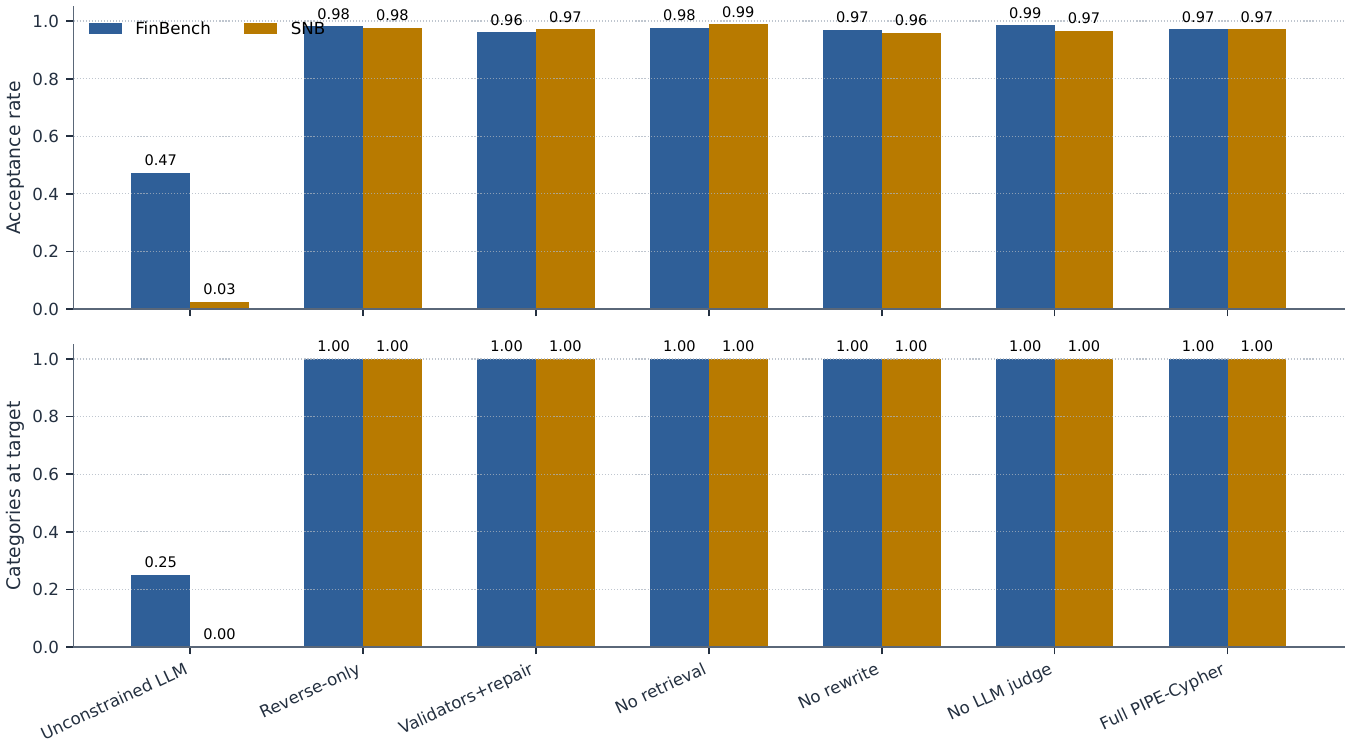}
\caption{Target-100 ablation yield on FinBench and SNB. Unconstrained local generation is reported as a stress baseline with explicit attempt accounting; the execution-grounded governed variants reach all eight workload-category targets on both graphs. This shows that the grounded governance core is reliable at filling the planned benchmark cells; it does not claim that each optional stage independently increases yield.}
\label{fig:ablation_suite_target100_main}
\end{figure}

Figure~\ref{fig:ablation_suite_target100_main} and Table~\ref{tab:ablation_results} show the target-100 stress baseline. The unconstrained rows are deliberately harsh: raw local-model generations may look plausible, but they do not produce balanced, executable benchmark coverage. The governed variants fill every target cell on both graphs.

\begin{table}[H]
\centering
\small
\resizebox{\columnwidth}{!}{%
\begin{tabular}{llrrrrr}
\toprule
Setting & Graph & Attempts & Records & Accepted & Acceptance & Categories at target \\
\midrule
Unconstrained LLM & FinBench & 422 & 422 & 200 & 0.474 & 2/8 \\
Unconstrained LLM & SNB & 2,000 & 2,000 & 50 & 0.025 & 0/8 \\
Reverse-only & FinBench & 815 & 815 & 800 & 0.982 & 8/8 \\
Reverse-only & SNB & 820 & 820 & 800 & 0.976 & 8/8 \\
Validators+repair & FinBench & 833 & 833 & 800 & 0.960 & 8/8 \\
Validators+repair & SNB & 824 & 824 & 800 & 0.971 & 8/8 \\
No retrieval & FinBench & 819 & 819 & 800 & 0.977 & 8/8 \\
No retrieval & SNB & 809 & 809 & 800 & 0.989 & 8/8 \\
No rewrite & FinBench & 826 & 826 & 800 & 0.969 & 8/8 \\
No rewrite & SNB & 834 & 834 & 800 & 0.959 & 8/8 \\
No LLM judge & FinBench & 812 & 812 & 800 & 0.985 & 8/8 \\
No LLM judge & SNB & 828 & 828 & 800 & 0.966 & 8/8 \\
Full PIPE-Cypher & FinBench & 824 & 824 & 800 & 0.971 & 8/8 \\
Full PIPE-Cypher & SNB & 824 & 824 & 800 & 0.971 & 8/8 \\
\bottomrule
\end{tabular}
}
\caption{Live target-100 ablation evidence with local Qwen3.5-9B. Governed graph runs target 100 accepted examples per category; the unconstrained row is a stress baseline reported with explicit attempt accounting.}
\label{tab:ablation_results}
\end{table}

\subsection{Gate-Level Quality}

\begin{figure}[H]
\centering
\includegraphics[width=\columnwidth]{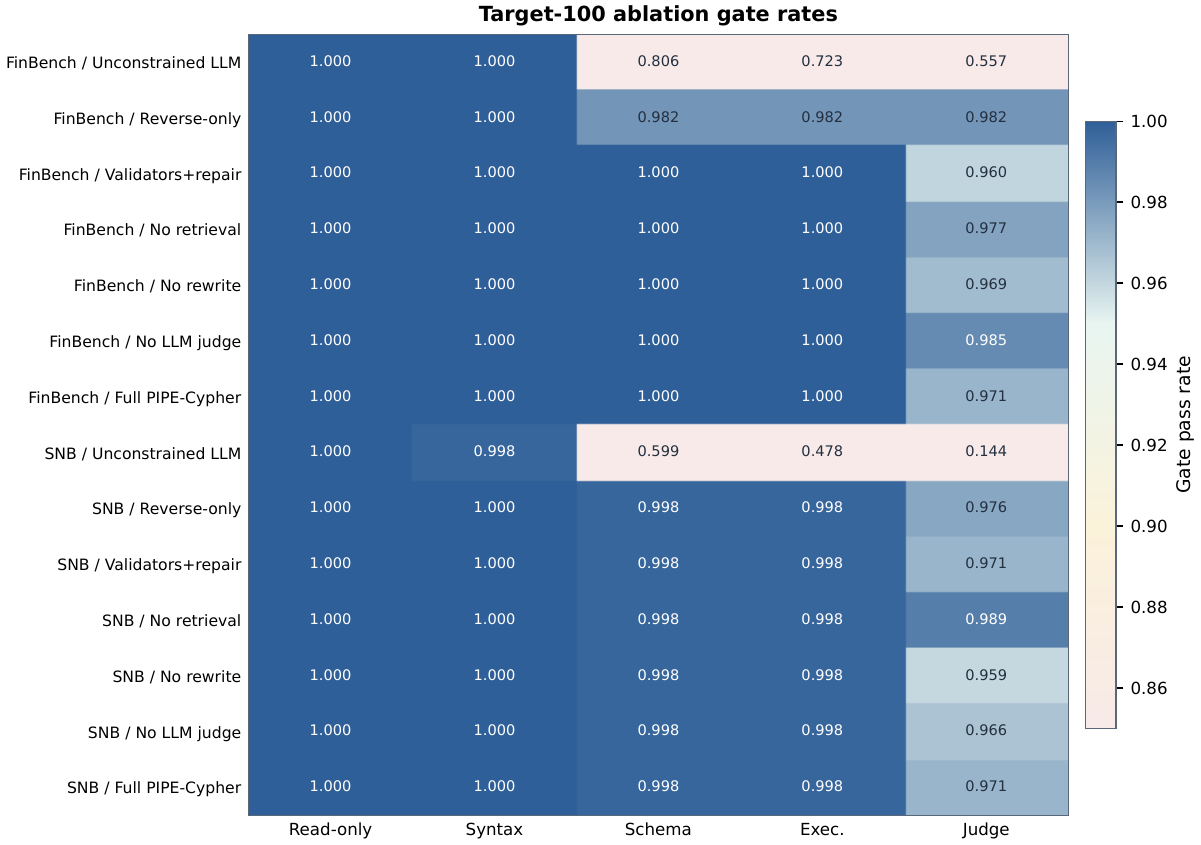}
\caption{Gate-rate heatmap for the audited target-100 ablation suite. Deterministic read-only and syntax gates are saturated; execution and judge rates expose the remaining quality differences across graph and pipeline variants.}
\label{fig:ablation_quality_target100}
\end{figure}

Yield alone is not enough. A benchmark factory must show which checks keep bad examples out. Table~\ref{tab:ablation_quality} and Figure~\ref{fig:ablation_quality_target100} report read-only, syntax, schema, execution, and judge/post-hoc rates for the same target-100 suite. Read-only and syntax checks saturate once governance is active; execution and semantic judging expose the remaining differences.

\begin{table}[H]
\centering
\small
\resizebox{\columnwidth}{!}{%
\begin{tabular}{llrrrrr}
\toprule
Setting & Graph & Read-only & Syntax & Schema & Exec. & Judge/post-hoc \\
\midrule
Unconstrained LLM & FinBench & 1.000 & 1.000 & 0.806 & 0.723 & 0.557 \\
Unconstrained LLM & SNB & 1.000 & 0.998 & 0.599 & 0.478 & 0.144 \\
Reverse-only & FinBench & 1.000 & 1.000 & 0.982 & 0.982 & 0.982 \\
Reverse-only & SNB & 1.000 & 1.000 & 0.998 & 0.998 & 0.976 \\
Validators+repair & FinBench & 1.000 & 1.000 & 1.000 & 1.000 & 0.960 \\
Validators+repair & SNB & 1.000 & 1.000 & 0.998 & 0.998 & 0.971 \\
No retrieval & FinBench & 1.000 & 1.000 & 1.000 & 1.000 & 0.977 \\
No retrieval & SNB & 1.000 & 1.000 & 0.998 & 0.998 & 0.989 \\
No rewrite & FinBench & 1.000 & 1.000 & 1.000 & 1.000 & 0.969 \\
No rewrite & SNB & 1.000 & 1.000 & 0.998 & 0.998 & 0.959 \\
No LLM judge & FinBench & 1.000 & 1.000 & 1.000 & 1.000 & 0.985 \\
No LLM judge & SNB & 1.000 & 1.000 & 0.998 & 0.998 & 0.966 \\
Full PIPE-Cypher & FinBench & 1.000 & 1.000 & 1.000 & 1.000 & 0.971 \\
Full PIPE-Cypher & SNB & 1.000 & 1.000 & 0.998 & 0.998 & 0.971 \\
\bottomrule
\end{tabular}
}
\caption{Quality-gate rates for the live target-100 ablation suite. Rates are computed over all generated records in each graph/setting; for no-judge settings, the judge column is a post-hoc scoring diagnostic.}
\label{tab:ablation_quality}
\end{table}

\subsection{Repeated-Suite Stability}

The same pattern holds across target sizes and seeds. The repeated-suite comparison normalizes by each suite's planned target, so it measures stability rather than rewarding larger raw counts. Full PIPE-Cypher and the governed ablations reach complete target coverage on both graphs, which is the behavior we want from a repeatable benchmark generator.

\begin{table}[H]
\centering
\small
\resizebox{\columnwidth}{!}{%
\begin{tabular}{llrrrrrr}
\toprule
Setting & Graph & Suites & Target cov. & Acceptance & Cat. target & Exec. & Judge \\
\midrule
No LLM judge & FinBench & 3/3 & 1.000 & 0.994$\pm$0.008 & 1.000 & 1.000 & 0.994 \\
No retrieval & FinBench & 3/3 & 1.000 & 0.967$\pm$0.031 & 1.000 & 1.000 & 0.967 \\
No rewrite & FinBench & 3/3 & 1.000 & 0.969$\pm$0.023 & 1.000 & 1.000 & 0.969 \\
Full PIPE-Cypher & FinBench & 3/3 & 1.000 & 0.975$\pm$0.016 & 1.000 & 1.000 & 0.975 \\
Reverse-only & FinBench & 3/3 & 1.000 & 0.994$\pm$0.011 & 1.000 & 0.994 & 0.994 \\
Validators+repair & FinBench & 3/3 & 1.000 & 0.987$\pm$0.023 & 1.000 & 1.000 & 0.987 \\
No LLM judge & SNB & 3/3 & 1.000 & 0.982$\pm$0.015 & 1.000 & 0.996 & 0.982 \\
No retrieval & SNB & 3/3 & 1.000 & 0.993$\pm$0.004 & 1.000 & 0.996 & 0.993 \\
No rewrite & SNB & 3/3 & 1.000 & 0.983$\pm$0.021 & 1.000 & 0.996 & 0.983 \\
Full PIPE-Cypher & SNB & 3/3 & 1.000 & 0.987$\pm$0.014 & 1.000 & 0.996 & 0.987 \\
Reverse-only & SNB & 3/3 & 1.000 & 0.989$\pm$0.011 & 1.000 & 0.996 & 0.989 \\
Validators+repair & SNB & 3/3 & 1.000 & 0.987$\pm$0.014 & 1.000 & 0.996 & 0.987 \\
\bottomrule
\end{tabular}
}
\caption{Target-size and repeated-seed ablation sensitivity. Target coverage normalizes accepted examples by each suite's planned graph/category target, so target-50 and target-100 suites can be compared without treating larger raw counts as quality gains. Unconstrained local generation is excluded from this stability table and reported separately as the attempt-logged stress baseline in Table~\ref{tab:ablation_results}.}
\label{tab:ablation_suite_comparison}
\end{table}

\begin{figure}[H]
\centering
\includegraphics[width=\columnwidth]{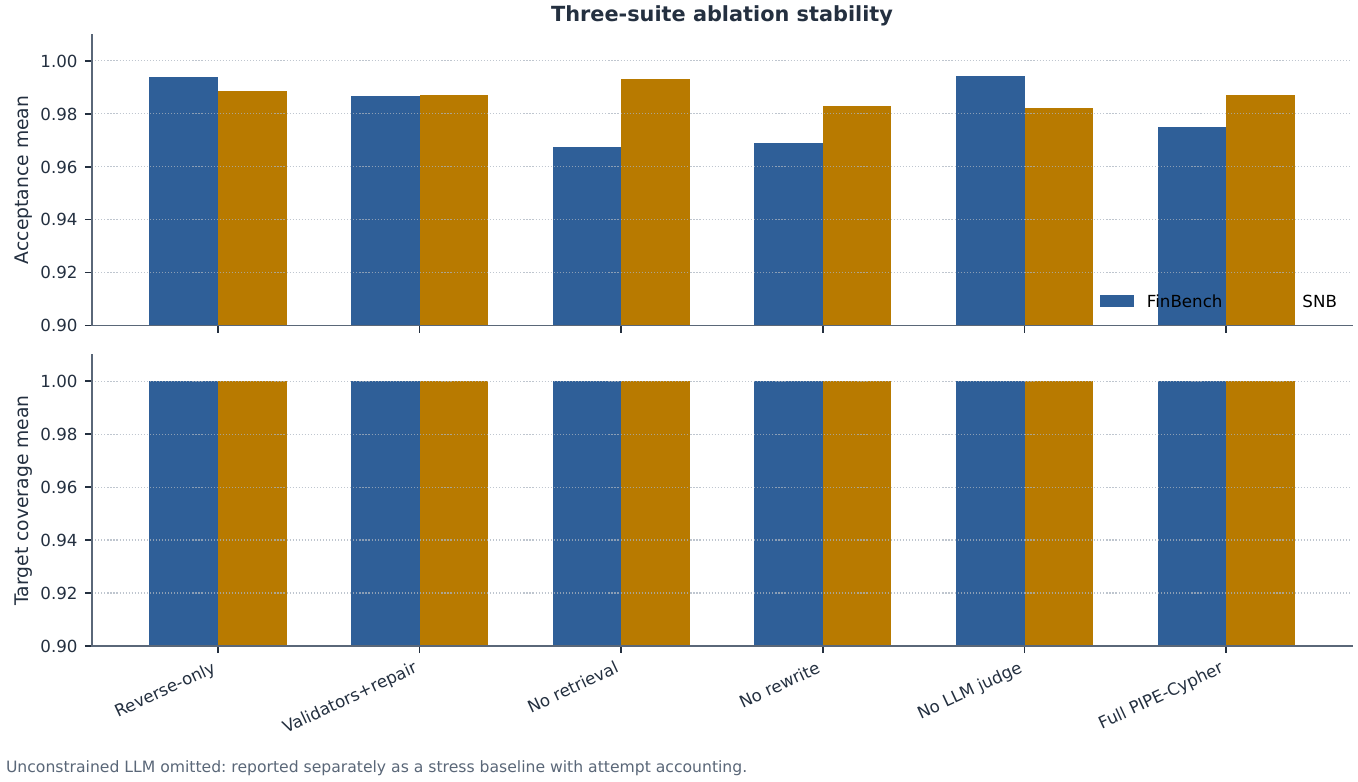}
\caption{Three-suite ablation stability over the original target-50 suite, corrected target-100 suite, and seed-17 target-50 repeat. Target-normalized coverage stays at 1.000 for all non-unconstrained cells.}
\label{fig:ablation_suite_comparison}
\end{figure}

\subsection{Rejected Candidate Taxonomy}

The rejection taxonomy shows where the remaining work occurs. After Cypher validation, schema-invalid candidates are rare. Most rejections are duplicate/diversity blocks from recovery or empty-result candidates caught before export. This is the behavior an enterprise deployment wants: invalid or brittle examples remain in the ledger, not in the benchmark.

\begin{table}[H]
\centering
\small
\resizebox{\columnwidth}{!}{%
\begin{tabular}{lrr}
\toprule
Failure bucket & Count & Share of rejected \\
\midrule
Diversity/duplicate control & 1,366 & 0.710 \\
Empty result & 486 & 0.252 \\
Judge semantic reject & 67 & 0.035 \\
Execution error & 4 & 0.002 \\
Schema invalid & 2 & 0.001 \\
\bottomrule
\end{tabular}
}
\caption{Failure taxonomy over full-run generation candidates before benchmark export. Accepted examples are excluded from the bucket shares.}
\label{tab:failure_taxonomy}
\end{table}

\begin{figure}[H]
\centering
\includegraphics[width=\columnwidth]{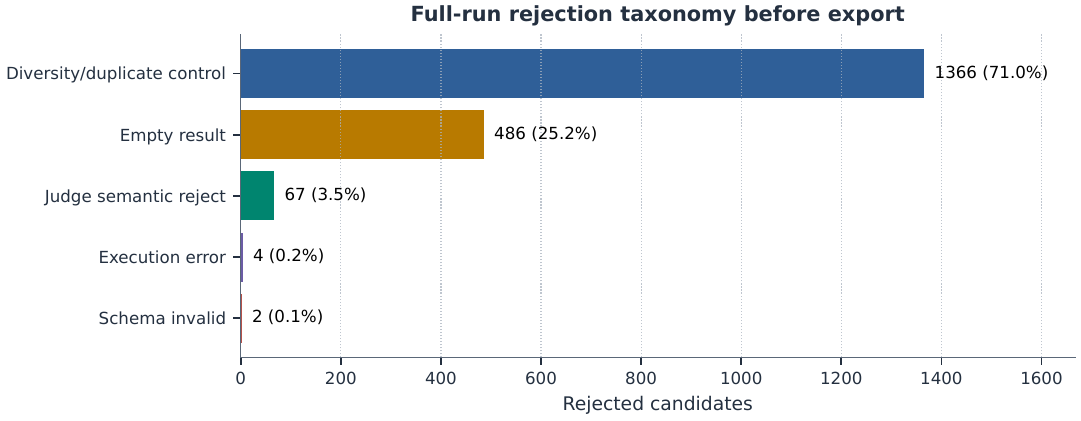}
\caption{Full-run rejection taxonomy before benchmark export. Most rejected candidates come from duplicate/diversity control during category recovery and from empty execution results; schema-invalid Cypher is rare after Cypher validation.}
\label{fig:failure_taxonomy}
\end{figure}

\FloatBarrier

\section{Benchmark Artifact and Third-Graph Onboarding}

\subsection{Export Balance}

\begin{figure}[H]
\centering
\includegraphics[width=\columnwidth]{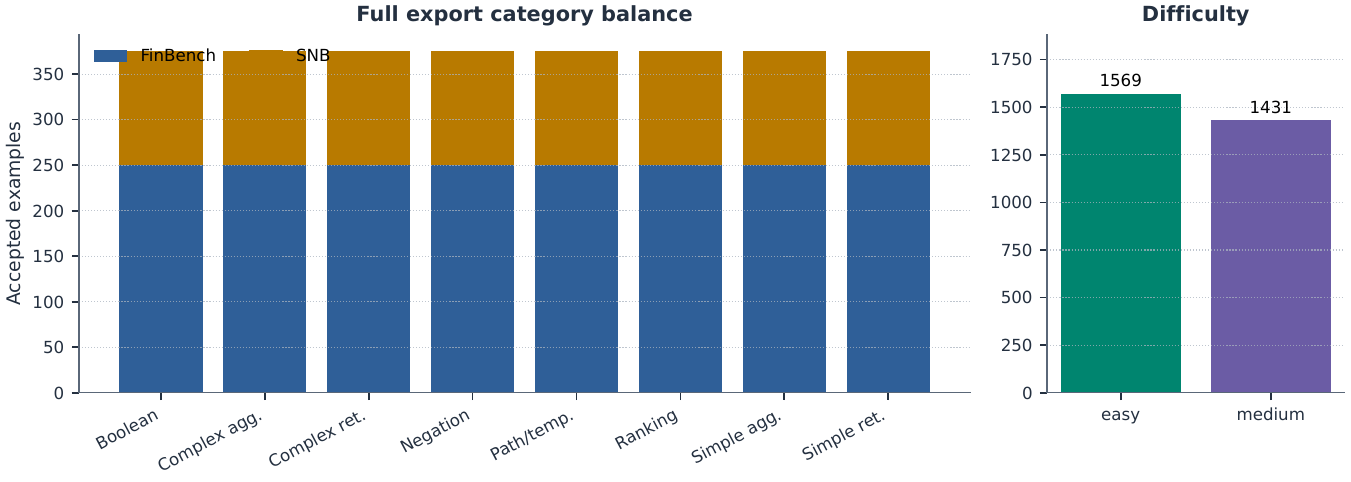}
\caption{Full 3,000-example export distribution. The benchmark preserves the planned 2:1 FinBench/SNB graph mix while balancing all eight Cypher workload categories and maintaining a near-even easy/medium difficulty split.}
\label{fig:full_export_distribution}
\end{figure}

The exported benchmark makes scale and balance visible. Figure~\ref{fig:full_export_distribution} shows the planned 2:1 FinBench/SNB mix, exact category balance, and near-even difficulty split. PIPE-Cypher produces a benchmark that is balanced by design rather than sampled opportunistically.

\subsection{Graph Scale and Schema Variety}

\begin{figure}[H]
\centering
\resizebox{\columnwidth}{!}{%
\begin{tikzpicture}[
  schemaNode/.style={draw=blue!55!black, rounded corners=2pt, fill=blue!6, thick, align=center, font=\footnotesize\bfseries, minimum width=2.0cm, minimum height=0.75cm},
  rel/.style={-Latex, thick, draw=blue!65!black},
  relLabel/.style={font=\scriptsize\ttfamily, align=center, fill=white, inner sep=1.4pt}
]
\node[schemaNode] (person) at (0,0) {Person\\{\scriptsize 8 props}};
\node[schemaNode] (company) at (0,-2.2) {Company\\{\scriptsize 9 props}};
\node[schemaNode] (account) at (4.0,-1.1) {Account\\{\scriptsize 9 props}};
\node[schemaNode] (loan) at (7.8,0) {Loan\\{\scriptsize 6 props}};
\node[schemaNode] (medium) at (7.8,-2.2) {Medium\\{\scriptsize 5 props}};

\draw[rel] (person) -- node[relLabel, above, sloped] {OWN\_ACCOUNT} (account);
\draw[rel] (company) -- node[relLabel, below, sloped] {OWN\_ACCOUNT} (account);
\draw[rel] (person) -- node[relLabel, above, sloped] {APPLY\_LOAN} (loan);
\draw[rel] (company) to[bend right=12] node[relLabel, below, sloped] {APPLY\_LOAN} (loan);
\draw[rel] (account) -- node[relLabel, above, sloped] {REPAY} (loan);
\draw[rel] (loan) -- node[relLabel, above, sloped] {DEPOSIT} (account);
\draw[rel] (medium) -- node[relLabel, below, sloped] {SIGN\_IN} (account);
\path[rel] (account) edge[loop right, min distance=15mm] node[relLabel, right] {TRANSFER\_TO\\WITHDRAW\_TO} (account);
\path[rel] (person) edge[loop left, min distance=14mm] node[relLabel, left] {GUARANTEE} (person);
\path[rel] (company) edge[loop left, min distance=14mm] node[relLabel, left] {INVEST\\GUARANTEE} (company);
\draw[rel] (person) to[bend right=13] node[relLabel, left] {INVEST} (company);
\end{tikzpicture}}
\caption{FinBench schema graph used in the reported runs. The workload is centered on financial entities, account ownership, transaction/event relationships, loans, sign-in media, guarantees, and investments.}
\label{fig:finbench_schema}
\end{figure}
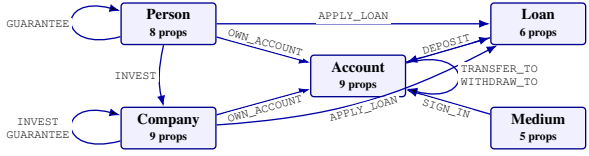

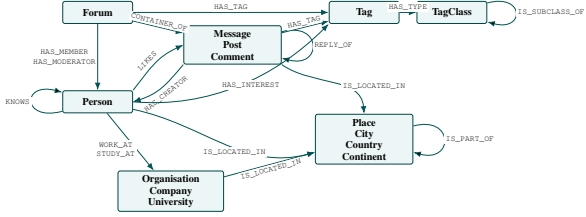
\begin{figure}[H]
\centering
\resizebox{\columnwidth}{!}{%
\begin{tikzpicture}[
  schemaNode/.style={draw=teal!55!black, rounded corners=2pt, fill=teal!7, thick, align=center, font=\footnotesize\bfseries, minimum width=1.85cm, minimum height=0.68cm},
  rel/.style={-Latex, thick, draw=teal!65!black},
  relLabel/.style={font=\scriptsize\ttfamily, align=center, fill=white, inner sep=1.4pt}
]
\node[schemaNode, minimum width=2.15cm] (person) at (0,0) {Person};
\node[schemaNode, minimum width=2.15cm] (forum) at (0,2.75) {Forum};
\node[schemaNode, minimum width=3.0cm] (content) at (4.15,1.75) {Message\\Post\\Comment};
\node[schemaNode, minimum width=2.15cm] (tag) at (8.2,2.75) {Tag};
\node[schemaNode, minimum width=2.15cm] (tagclass) at (10.9,2.75) {TagClass};
\node[schemaNode, minimum width=3.0cm] (place) at (8.2,-1.15) {Place\\City\\Country\\Continent};
\node[schemaNode, minimum width=3.25cm] (org) at (2.25,-2.75) {Organisation\\Company\\University};

\draw[rel] (forum) -- node[relLabel, above, sloped] {CONTAINER\_OF} (content);
\draw[rel] (forum) -- node[relLabel, left] {HAS\_MEMBER\\HAS\_MODERATOR} (person);
\draw[rel] (forum) -- node[relLabel, above, sloped] {HAS\_TAG} (tag);
\draw[rel] (content.south west) to[out=230,in=18,looseness=1.10] node[relLabel, below, sloped, pos=0.50, yshift=-2.5pt] {HAS\_CREATOR} (person.east);
\draw[rel] (person.north east) to[out=52,in=205,looseness=1.10] node[relLabel, above, sloped, pos=0.44, yshift=2.5pt] {LIKES} (content.west);
\path[rel] (content) edge[loop right, min distance=13mm] node[relLabel, right] {REPLY\_OF} (content);
\draw[rel] (content) -- node[relLabel, above, sloped] {HAS\_TAG} (tag);
\draw[rel] (person.east) to[out=-5,in=-135,looseness=0.92] node[relLabel, below, pos=0.56] {HAS\_INTEREST} (tag.south west);
\draw[rel] (tag) -- node[relLabel, above] {HAS\_TYPE} (tagclass);
\path[rel] (tagclass) edge[loop right, min distance=12mm] node[relLabel, right] {IS\_SUBCLASS\_OF} (tagclass);
\path[rel] (person) edge[loop left, min distance=13mm] node[relLabel, left] {KNOWS} (person);
\draw[rel] (person) -- node[relLabel, left, pos=0.64] {WORK\_AT\\STUDY\_AT} (org);
\draw[rel] (person) to[out=-12,in=195] node[relLabel, below, pos=0.56] {IS\_LOCATED\_IN} (place);
\draw[rel] (content.south east) to[out=-35,in=95] node[relLabel, right, pos=0.58] {IS\_LOCATED\_IN} (place.north);
\draw[rel] (org) -- node[relLabel, below, sloped] {IS\_LOCATED\_IN} (place);
\path[rel] (place) edge[loop right, min distance=13mm] node[relLabel, right] {IS\_PART\_OF} (place);
\end{tikzpicture}}
\caption{SNB schema graph used in the reported runs. Related labels are grouped into content, place, and organization families so all 14 labels and 59 directed label/type/label patterns remain readable; Table~\ref{tab:graph_statistics} gives the full inventory counts.}
\label{fig:snb_schema}
\end{figure}

\begin{figure}[H]
\centering
\resizebox{\columnwidth}{!}{%
\begin{tikzpicture}[
  schemaNode/.style={draw=violet!55!black, rounded corners=2pt, fill=violet!7, thick, align=center, font=\footnotesize\bfseries, minimum width=2.2cm, minimum height=0.76cm},
  rel/.style={-Latex, thick, draw=violet!65!black},
  relLabel/.style={font=\tiny\ttfamily, align=center, fill=white, inner sep=1.1pt}
]
\node[schemaNode] (officer) at (0,1.85) {Officer\\{\scriptsize 19 props}};
\node[schemaNode] (intermediary) at (0,-1.85) {Intermediary\\{\scriptsize 19 props}};
\node[schemaNode] (entity) at (5.15,0) {Entity\\{\scriptsize 29 props}};
\node[schemaNode] (address) at (9.85,0.65) {Address\\{\scriptsize 15 props}};
\node[schemaNode] (other) at (5.15,-3.35) {Other\\{\scriptsize 14 props}};

\draw[rel] (officer.east) to[bend left=12] node[relLabel, above, sloped, pos=0.22, yshift=3pt] {officer\_of\\intermediary\_of\\connected\_to} (entity.north west);
\draw[rel] (intermediary.east) to[bend right=10] node[relLabel, below, sloped, pos=0.36] {intermediary\_of\\officer\_of\\connected\_to} (entity.south west);
\draw[rel] (officer.north east) to[out=28,in=165] node[relLabel, above, pos=0.64] {registered\_address} (address.north west);
\draw[rel] (intermediary.south east) to[out=-18,in=-155] node[relLabel, below, pos=0.56] {registered\_address} (address.south west);
\draw[rel] (entity) -- node[relLabel, above, pos=0.72] {registered\_address} (address);
\draw[rel] (other.east) to[out=-5,in=-112] node[relLabel, below, sloped, pos=0.58] {registered\_address} (address.south);
\draw[rel] (other.north) -- node[relLabel, left, pos=0.70, xshift=-4pt, yshift=-2pt] {connected\_to\\underlying} (entity.south);
\draw[rel] (officer.west) -- (-1.85,1.85) -- (-1.85,-3.35) -- (other.west);
\node[relLabel, anchor=east] at (-1.92,-2.72) {officer\_of};
\path[rel] (officer) edge[loop left, min distance=15mm] node[relLabel, left] {same\_name\_as\\same\_id\_as\\similar\\underlying} (officer);
\path[rel] (entity) edge[loop above, min distance=16mm] node[relLabel, above] {same\_company\_as\\same\_name\_as\\same\_as\\similar\_company\_as} (entity);
\path[rel] (intermediary) edge[loop left, min distance=14mm] node[relLabel, left] {same\_name\_as\\similar\\same\_intermediary\_as} (intermediary);
\path[rel] (address) edge[loop right, min distance=13mm] node[relLabel, right] {same\_as\\same\_address\_as} (address);
\draw[rel] (entity.north west) to[out=152,in=-8,looseness=0.9] node[relLabel, below, sloped, pos=0.52, yshift=-6pt] {same\_name\_as} (officer.east);
\draw[rel] (intermediary.north west) to[out=135,in=-132] node[relLabel, left, pos=0.56] {same\_as\\underlying\\similar} (officer.south west);
\end{tikzpicture}}
\caption{ICIJ Offshore Leaks schema graph used for the third-graph onboarding audit. The schema is compact in node labels but large in edge volume: relationship families encode officer/entity/intermediary roles, registered addresses, identity-resolution links, and similarity links.}
\label{fig:icij_schema}
\end{figure}
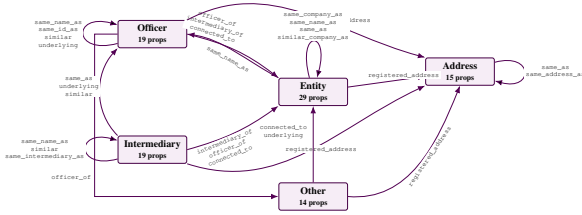

Table~\ref{tab:graph_statistics} summarizes the graph sizes and live schema inventories before the third-graph onboarding audit. FinBench and SNB are the controlled LDBC workloads. ICIJ is the public finance/compliance graph we use to test whether the same onboarding code works outside the two original schemas. Figures~\ref{fig:finbench_schema}--\ref{fig:icij_schema} show the same schemas in graph form. Node boxes are labels; arrows are directed relationship families observed by introspection. Some arrows group repeated label-pair alternatives so the figure remains readable, while Table~\ref{tab:graph_statistics} gives the full inventory counts.

\begin{table}[H]
\centering
\footnotesize
\resizebox{\columnwidth}{!}{%
\begin{tabular}{lrrrrrrr}
\toprule
Graph & Nodes & Edges & Labels & Rel. types & Patterns & Node props & Rel. props \\
\midrule
FinBench & 10,006 & 57,622 & 5 & 9 & 13 & 37 & 32 \\
SNB & 34,735 & 70,842 & 14 & 15 & 59 & 62 & 5 \\
ICIJ Offshore Leaks & 2,016,523 & 3,339,267 & 5 & 14 & 64 & 82 & 48 \\
\bottomrule
\end{tabular}
}
\caption{Study graph size and schema inventory from live introspection. Patterns are directed start-label/type/end-label triples; property counts are distinct label-property or relationship-type-property fields. ICIJ Offshore Leaks is used as a public third-graph onboarding audit beyond the two LDBC workloads.}
\label{tab:graph_statistics}
\end{table}

\FloatBarrier

\subsection{ICIJ Onboarding}

\begin{figure}[H]
\centering
\includegraphics[width=\columnwidth]{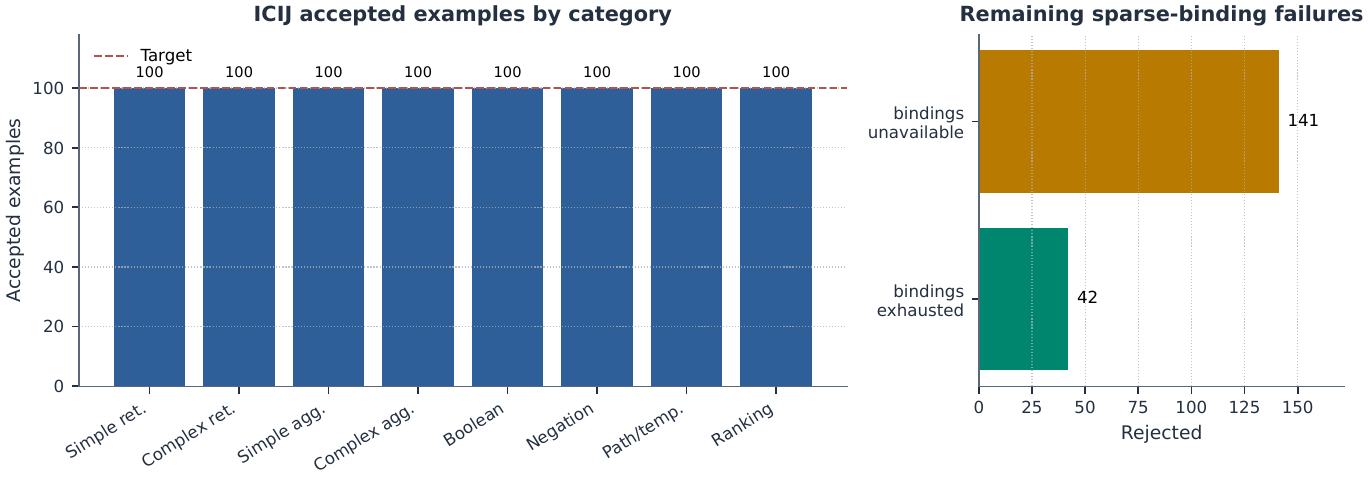}
\caption{ICIJ Offshore Leaks onboarding audit. Schema-derived sparse-category templates recover balanced target-100 coverage on a public finance/compliance graph beyond the two LDBC workloads.}
\label{fig:icij_onboarding_audit}
\end{figure}

Table~\ref{tab:icij_onboarding} and Figure~\ref{fig:icij_onboarding_audit} report the third-graph result. ICIJ matters because it forced PIPE-Cypher to derive sparse-category templates from the schema instead of relying on preauthored FinBench/SNB templates. The run reaches all eight category targets while using the same validation and audit standards.

\begin{table}[H]
\centering
\small
\resizebox{\columnwidth}{!}{%
\begin{tabular}{lr}
\toprule
ICIJ onboarding property & Value \\
\midrule
Graph nodes / relationships & 2,016,523 / 3,339,267 \\
Labels / relationship types & 5 / 14 \\
Generated / accepted & 983 / 800 \\
Acceptance rate & 0.814 \\
Categories at target & 8/8 \\
Study audit & ready \\
Sparse schema-derived accepts & complex agg. 97, negation 28, ranking 98 \\
\bottomrule
\end{tabular}
}
\caption{ICIJ Offshore Leaks third-graph onboarding audit. The public finance/compliance graph tests arbitrary-schema generation beyond the two LDBC study workloads; raw values remain outside the reported artifacts.}
\label{tab:icij_onboarding}
\end{table}

\subsection{Category Crosswalk}

Table~\ref{tab:category_crosswalk} connects PIPE-Cypher's eight categories to the simpler retrieval/aggregation/evaluation-query taxonomy used in Mind the Query. The crosswalk keeps the comparison clear while making our extra enterprise workload classes explicit: negation, temporal/path reasoning, and ranking/top-k.

\begin{table}[H]
\centering
\small
\resizebox{\columnwidth}{!}{%
\begin{tabular}{lll}
\toprule
PIPE-Cypher category & Closest MTQ category & Added enterprise distinction \\
\midrule
Simple retrieval & SR & Direct node/edge lookup with exact filters. \\
Complex retrieval & CR & Multi-hop or multi-pattern retrieval. \\
Simple aggregation & SA & Single aggregation such as count/min/max/average. \\
Complex aggregation & CA & Grouped or multi-stage aggregation over graph neighborhoods. \\
Boolean existence & EQ & Precise yes/no or existence answer. \\
Negation/difference & CR & Absence, anti-join, or difference query. \\
Path/temporal transaction & CR/CA & Temporal or path-oriented transaction neighborhood. \\
Ranking/top-k & SA/CA & Ordered top-k query with explicit limit. \\
\bottomrule
\end{tabular}
}
\caption{Category crosswalk to Mind the Query. PIPE-Cypher keeps the familiar retrieval/aggregation/evaluation-query structure while adding enterprise workloads such as negation, temporal paths, and ranking.}
\label{tab:category_crosswalk}
\end{table}

\FloatBarrier

\section{Downstream Transfer and Example-Bank Utility}

\subsection{Transfer Controls}

The downstream experiment tests whether the benchmark is useful for model evaluation. A useful enterprise benchmark should not merely reward syntactically valid Cypher. It should reveal when a model produces a query that parses, mentions plausible schema elements, and even executes, but answers the wrong operational question. The Qwen3.5-9B result has exactly that shape: parse validity is 0.963 and schema validity is 0.916, while exact execution accuracy is only 0.189 on the full held-out split. The gap shows that PIPE-Cypher measures semantic graph-query competence rather than only query formatting.

\begin{figure*}[!htbp]
\centering
\includegraphics[width=\textwidth]{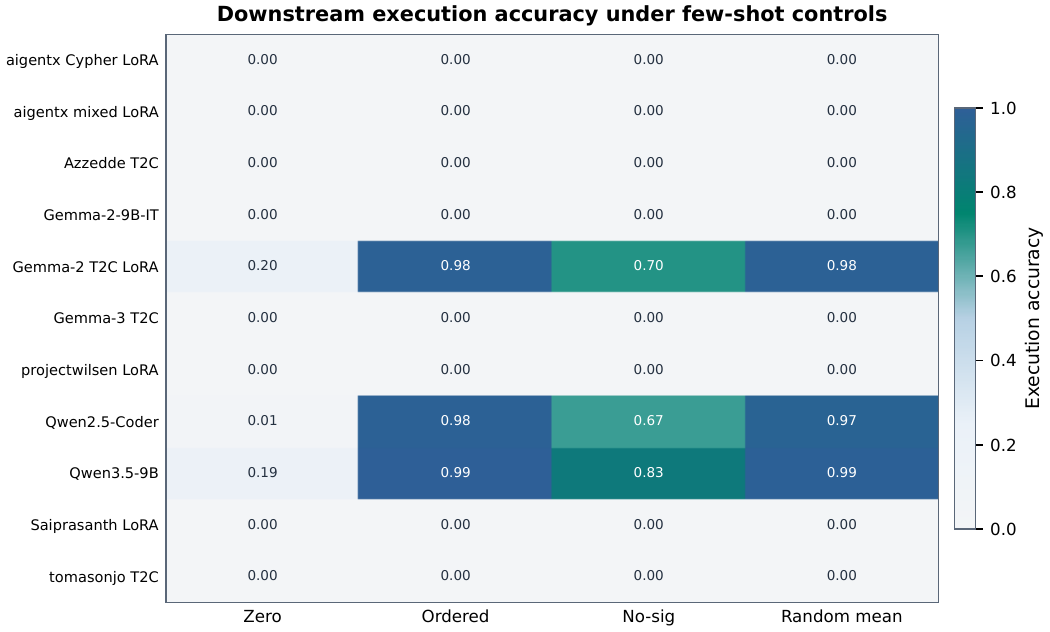}
\caption{Execution accuracy for 11 completed local downstream models under zero-shot and few-shot control modes. The heatmap highlights both findings: schema-specific examples can sharply help compatible models, and several public fine-tuned checkpoints still fail under enterprise-style Cypher prompting.}
\label{fig:downstream_fewshot_controls}
\end{figure*}

The multi-model transfer suite compares local general instruction, code-tuned, Cypher instruction, and Text2Cypher-finetuned models while keeping the schema prompt, evaluation backend, split, and execution metrics fixed. The exact checkpoint and adapter provenance is listed in Table~\ref{tab:downstream_model_provenance}. Figure~\ref{fig:downstream_fewshot_controls} shows a sharp pattern: zero-shot transfer is weak, but accepted graph-specific examples can become a useful private question-answer bank. Demonstration-bank controls close much of the gap for compatible model families such as Qwen, Qwen-Coder, and one Gemma Text2Cypher LoRA. Several public fine-tuned checkpoints remain brittle under enterprise-style prompting.

\begin{table*}[t]
\centering
\scriptsize
\resizebox{\textwidth}{!}{%
\begin{tabular}{@{}llll@{}}
\toprule
Displayed model & Served checkpoint or adapter & Family & Citation \\
\midrule
aigentx/Llama-3.1-8B Cypher LoRA & \texttt{aigentx/llama-3.1-8b-instruct-cypher} & Cypher LoRA & \citep{llama3herd2024,aigentx2025llama31cypherhf} \\
aigentx/Llama-3.1-8B Cypher mixed LoRA & \texttt{aigentx/llama-3.1-8b-instruct-cypher-mixed-samples} & Cypher mixed LoRA & \citep{llama3herd2024,aigentx2025llama31cyphermixedhf} \\
Azzedde/llama3.1-8b-text2cypher & \texttt{Azzedde/llama3.1-8b-text2cypher} & Text2Cypher fine-tuned & \citep{llama3herd2024,azzedde2025llama31text2cypherhf} \\
Gemma-2-9B-IT & \texttt{google/gemma-2-9b-it} & General instruction & \citep{gemma2team2024} \\
neo4j/Gemma-2-9B Text2Cypher LoRA & \texttt{neo4j/text2cypher-gemma-2-9b-it-finetuned-2024v1} & Text2Cypher LoRA & \citep{gemma2team2024,text2cypher2025,neo4j2025gemma2text2cypherhf} \\
neo4j/Gemma-3-4B Text2Cypher & \texttt{neo4j/text-to-cypher-Gemma-3-4B-Instruct-2025.04.0} & Text2Cypher fine-tuned & \citep{gemma3team2025,text2cypher2025,neo4j2025gemma3text2cypherhf} \\
projectwilsen/Llama-3.1-8B Text2Cypher LoRA & \texttt{projectwilsen/llama3.1-8b-text2cypher-neo4j-live} & Text2Cypher LoRA & \citep{llama3herd2024,projectwilsen2024llama31text2cypherhf} \\
Qwen2.5-Coder-7B-Instruct & \texttt{Qwen/Qwen2.5-Coder-7B-Instruct} & Code instruction & \citep{hui2024qwen25coder} \\
Qwen3.5-9B & \texttt{Qwen/Qwen3.5-9B} & General instruction & \citep{yang2025qwen3,qwen2026qwen35hf} \\
Saiprasanth15/Llama-3.1-8B Text2Cypher LoRA & \texttt{Saiprasanth15/llama3.1-8b-text2cypher-neo4j-live} & Text2Cypher LoRA & \citep{llama3herd2024,saiprasanth2024llama31text2cypherhf} \\
tomasonjo/text2cypher-demo-16bit & \texttt{tomasonjo/text2cypher-demo-16bit} & Text2Cypher fine-tuned & \citep{llama3herd2024,tomasonjo2024text2cypherdemohf} \\
\bottomrule
\end{tabular}
}
\caption{Downstream model provenance for the 11 completed local checkpoints in the transfer study. Original model-family papers are cited wherever available; Hugging Face repository citations are retained only to identify exact fine-tuned checkpoints or adapters without a separate paper.}
\label{tab:downstream_model_provenance}
\end{table*}

\begin{table*}[t]
\centering
\small
\resizebox{\textwidth}{!}{%
\begin{tabular}{llrrrrrr}
\toprule
Model & Tuning & Zero & Ordered & No-sig & Random $\mu$ & Random $\sigma$ & Best \\
\midrule
aigentx/Llama-3.1-8B Cypher LoRA & Cypher LoRA & 0.000 & 0.000 & 0.000 & 0.000 & 0.000 & no gain \\
aigentx/Llama-3.1-8B Cypher mixed LoRA & Cypher mixed LoRA & 0.000 & 0.000 & 0.000 & 0.000 & 0.000 & no gain \\
Azzedde/llama3.1-8b-text2cypher & Text2Cypher fine-tuned & 0.000 & 0.000 & 0.000 & 0.000 & 0.000 & no gain \\
Gemma-2-9B-IT & general instruction & 0.000 & 0.000 & 0.000 & 0.000 & 0.000 & no gain \\
neo4j/Gemma-2-9B Text2Cypher LoRA & Text2Cypher LoRA & 0.203 & 0.983 & 0.699 & 0.981 & 0.009 & ordered (0.983) \\
neo4j/Gemma-3-4B Text2Cypher & Text2Cypher fine-tuned & 0.000 & 0.000 & 0.000 & 0.000 & 0.000 & no gain \\
projectwilsen/Llama-3.1-8B Text2Cypher LoRA & Text2Cypher LoRA & 0.000 & 0.000 & 0.000 & 0.000 & 0.000 & no gain \\
Qwen2.5-Coder-7B-Instruct & code instruction & 0.007 & 0.983 & 0.669 & 0.972 & 0.002 & ordered (0.983) \\
Qwen3.5-9B & general instruction & 0.189 & 0.993 & 0.828 & 0.986 & 0.007 & ordered (0.993) \\
Saiprasanth15/Llama-3.1-8B Text2Cypher LoRA & Text2Cypher LoRA & 0.000 & 0.000 & 0.000 & 0.000 & 0.000 & no gain \\
tomasonjo/text2cypher-demo-16bit & Text2Cypher fine-tuned & 0.000 & 0.000 & 0.000 & 0.000 & 0.000 & no gain \\
\bottomrule
\end{tabular}
}
\caption{Few-shot demonstration-bank controls for local downstream Text2Cypher evaluation over completed 296-example local-model runs. Ordered uses the deterministic same-graph, same-category example bank; scored excludes exact query-signature matches and near-duplicate questions; random reports the mean and standard deviation across seeds 13, 17, and 23. ``No gain'' means no few-shot control exceeded that model's zero-shot execution accuracy. Model-family papers and exact checkpoint sources are listed in Table~\ref{tab:downstream_model_provenance}.}
\label{tab:downstream_fewshot_controls}
\end{table*}

Table~\ref{tab:downstream_fewshot_control_uncertainty} reports checkpoint-level uncertainty for the same controls. The interval is deliberately conservative because the unit is model checkpoint rather than question row; it supports the narrower claim that example-bank gains are model-family dependent, not universal.

\begin{table}[H]
\centering
\small
\resizebox{\columnwidth}{!}{%
\begin{tabular}{lrrrr}
\toprule
Control & Mean acc. & $\Delta$ vs zero & 95\% $\Delta$ CI & Models improved \\
\midrule
Ordered same-category & 0.269 & 0.233 & [0.000, 0.481] & 3/11 \\
Scored no-signature & 0.200 & 0.163 & [0.000, 0.335] & 3/11 \\
Random same-category mean & 0.267 & 0.231 & [0.000, 0.464] & 3/11 \\
\bottomrule
\end{tabular}
}
\caption{Model-level paired bootstrap uncertainty for downstream few-shot controls. The unit of resampling is the local checkpoint, not an individual question, so the interval is a conservative check on whether gains are broad across model families. Zero-shot mean execution accuracy is 0.036.}
\label{tab:downstream_fewshot_control_uncertainty}
\end{table}

Table~\ref{tab:fewshot_leakage_controls} gives the interpretation for the transfer result. Ordered and random same-category demonstrations are useful example-bank conditions. The scored no-signature condition is the stricter leakage-aware control.

\begin{table}[H]
\centering
\small
\resizebox{\columnwidth}{!}{%
\begin{tabular}{lrrrrr}
\toprule
Selection mode & Rows & Demos & Sig. match & High sim. & Mean sim. \\
\midrule
Ordered & 296 & 1480 & 0.866 & 0.389 & 0.825 \\
No-signature & 296 & 1480 & 0.000 & 0.000 & 0.585 \\
Random & 296 & 1480 & 0.854 & 0.409 & 0.824 \\
\bottomrule
\end{tabular}
}
\caption{Few-shot leakage controls for downstream demonstration-bank evaluation. The held-out split has 0 exact train/test question overlaps and 289 train/test query-signature overlaps; selection-mode rates show how often retrieved demonstrations share the test query signature or exceed the 0.90 normalized-question similarity threshold.}
\label{tab:fewshot_leakage_controls}
\end{table}

\FloatBarrier

\subsection{Downstream Failure Modes}

\begin{center}
\begin{minipage}{\columnwidth}
\centering
\small
\resizebox{\columnwidth}{!}{%
\begin{tabular}{lrr}
\toprule
Downstream outcome & Count & Share of incorrect \\
\midrule
Answer mismatch & 125 & 0.521 \\
Execution failed & 79 & 0.329 \\
Schema invalid & 25 & 0.104 \\
Parse invalid & 11 & 0.046 \\
\bottomrule
\end{tabular}
}
\captionsetup{hypcap=false}
\captionof{table}{Downstream Text2Cypher failure taxonomy for local Qwen3.5-9B on the full exported test split. Shares exclude exact-answer matches.}
\label{tab:downstream_error_taxonomy}
\end{minipage}
\end{center}

Table~\ref{tab:downstream_error_taxonomy} makes the downstream study actionable. An answer mismatch means the model produced executable Cypher for the wrong semantics, so better graph-specific examples, retrieval, or adaptation are the likely fixes. Execution failures point instead to unsupported operators, brittle literals, or missing repair rules. Parse and schema failures are the errors deterministic guards should catch before a query ever reaches users.

This is why the taxonomy appears next to the transfer results. Many incorrect outputs are not malformed queries; they run and return the wrong answer set. A syntax-only or schema-only benchmark would miss that failure. For a benchmark owner, the taxonomy is therefore a debugging interface: it shows whether the next improvement should target schema retrieval, prompt constraints, value grounding, operator repair, or tenant-specific adaptation.

\section{Diversity and Cypher Strategy Audit}

\subsection{Diversity Diagnostics}

\begin{figure}[H]
\centering
\includegraphics[width=\columnwidth]{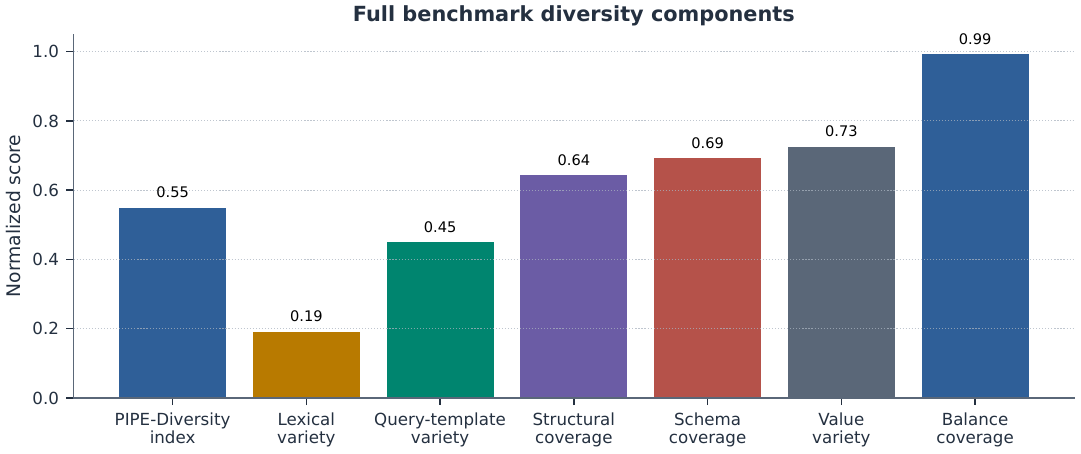}
\caption{Diversity diagnostics for the full 3,000-example export. Category, graph-category, and difficulty balance are strong by construction; schema coverage and query-signature diversity expose remaining concentration from the seeded-template run.}
\label{fig:diversity_diagnostics}
\end{figure}

Aggregate acceptance rates hide too much. The diversity and strategy diagnostics show whether the benchmark is balanced, which Cypher operators it exercises, and where residual concentration remains. Figure~\ref{fig:diversity_diagnostics} shows the useful diversity picture: category, graph-category, and difficulty balance are strong by construction, while schema and value coverage remain quantities to monitor during refresh.

\begin{table}[!htbp]
\centering
\small
\resizebox{\columnwidth}{!}{%
\begin{tabular}{lr}
\toprule
Metric & Value \\
\midrule
PIPE-Diversity index & 0.549 \\
Question Distinct-1 & 0.039 \\
Question Distinct-2 & 0.085 \\
Question adjusted Distinct-2 & 0.266 \\
Question self-BLEU-2 (sampled) & 0.813 \\
Mean nearest-neighbor question Jaccard & 0.775 \\
Unique query-signature ratio & 0.041 \\
Top query-signature share & 0.083 \\
Template-family entropy & 0.826 \\
Operator-combination entropy & 0.880 \\
Unique structural substructures & 134 \\
Category normalized entropy & 1.000 \\
Graph-category normalized entropy & 0.980 \\
Difficulty normalized entropy & 0.998 \\
Label coverage & 0.941 \\
Relationship-type coverage & 0.708 \\
Property-name coverage & 0.426 \\
Unique grounded-value ratio & 0.357 \\
Grounded values exactly quoted & 0.823 \\
Aggregation / negation / ordering rates & 0.500 / 0.125 / 0.125 \\
\bottomrule
\end{tabular}
}
\caption{Diversity diagnostics for the full exported benchmark. PIPE-Diversity is a geometric mean of lexical, query-template, structural, schema, value, and balance components; component rows are shown so the composite score does not hide residual concentration.}
\label{tab:diversity_metrics}
\end{table}

\subsection{Diversity-Governed Selection}

Table~\ref{tab:diversity_improvement} reports the first diversity-improvement pass. We select a target-50-per-graph/category subset from the 3,000 accepted examples and compare it with a hash-balanced random subset of the same size. The selector uses an MMR-style novelty score over query signatures, template families, structural substructures, schema atoms, entity values, and question tokens after all quality gates have passed. It improves the PIPE-Diversity index, unique query-signature ratio, unique structural substructures, adjusted Distinct-2, and property coverage while preserving a signature-disjoint 640/80/80 split. The mixed template-family entropy and self-BLEU result is useful rather than embarrassing: it shows that post-hoc selection can improve coverage, but stronger diversity requires oversampling or inducing more source templates during generation when a graph/category cell has only one or two viable signatures.

\begin{table}[H]
\centering
\small
\resizebox{\columnwidth}{!}{%
\begin{tabular}{lrrr}
\toprule
Metric & Random balanced & Diversity governed & $\Delta$ \\
\midrule
PIPE-Diversity index & 0.557 & 0.575 & +0.017 \\
Unique query-signature ratio & 0.062 & 0.135 & +0.073 \\
Top signature share (lower better) & 0.062 & 0.062 & +0.000 \\
Template-family entropy & 0.903 & 0.868 & -0.035 \\
Operator-combination entropy & 0.901 & 0.911 & +0.010 \\
Unique structural substructures & 97.000 & 134.000 & +37.000 \\
Question self-BLEU-2 (lower better) & 0.850 & 0.866 & +0.016 \\
Adjusted Distinct-2 & 0.266 & 0.284 & +0.018 \\
Property coverage & 0.407 & 0.426 & +0.019 \\
\bottomrule
\end{tabular}
}
\caption{Balanced subset comparison at the same graph/category target. The diversity-governed selector applies MMR-style novelty over Cypher signatures, template families, structural substructures, schema atoms, values, and question tokens after quality gates have already passed; structural/schema gains are reported alongside residual template concentration.}
\label{tab:diversity_improvement}
\end{table}

\subsection{Cypher Strategy Coverage}

Strategy diagnostics complement category balance. Two questions can share a category while exercising different Cypher behavior. Conversely, a category-balanced dataset can still miss joins, paths, negation, ranking, or bounded-result patterns. The strategy matrix and downstream strategy outcomes show what the benchmark actually exercises.

\begin{table}[H]
\centering
\small
\resizebox{\columnwidth}{!}{%
\begin{tabular}{lrrrrr}
\toprule
Primary strategy & Examples & Share & Rel. patterns & Return arity & Downstream exec. acc. \\
\midrule
Aggregation & 1125 & 0.375 & 1.42 & 1.00 & 0.396 \\
Join-heavy & 375 & 0.125 & 2.33 & 2.33 & 0.000 \\
Negation & 375 & 0.125 & 1.89 & 2.84 & 0.000 \\
Order/rank & 375 & 0.125 & 1.87 & 3.41 & 0.000 \\
Path & 375 & 0.125 & 1.37 & 3.00 & 0.000 \\
Single hop & 375 & 0.125 & 1.00 & 2.33 & 0.324 \\
\bottomrule
\end{tabular}
}
\caption{Cypher strategy diagnostics over the full 3,000-example export. Strategy tags are derived from generated Cypher structure rather than from category labels; downstream execution accuracy is reported on the full held-out test split when a strategy appears there.}
\label{tab:strategy_diagnostics}
\end{table}

\begin{figure*}[!htbp]
\centering
\includegraphics[width=0.92\textwidth]{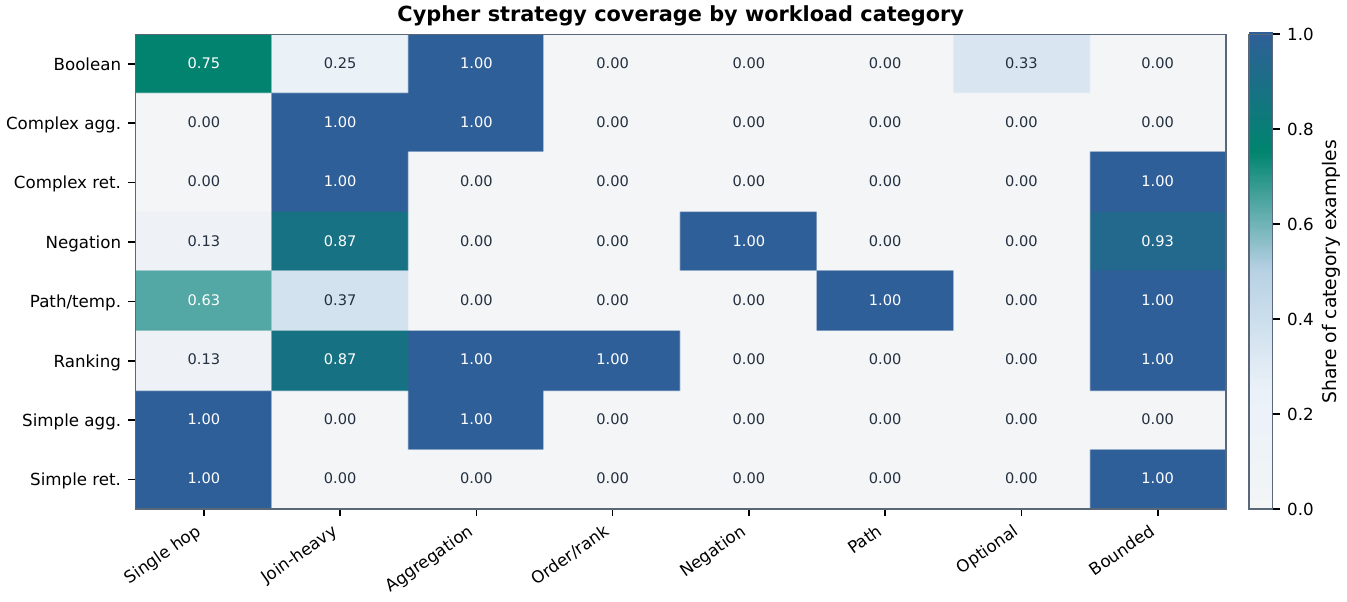}
\caption{Cypher strategy coverage by workload category. Category balancing does not by itself guarantee operator coverage; the strategy matrix shows where the benchmark exercises single-hop retrieval, joins, aggregation, ordering, negation, path patterns, optional matches, and bounded-result queries.}
\label{fig:strategy_coverage}
\end{figure*}

\begin{figure*}[!htbp]
\centering
\includegraphics[width=0.88\textwidth]{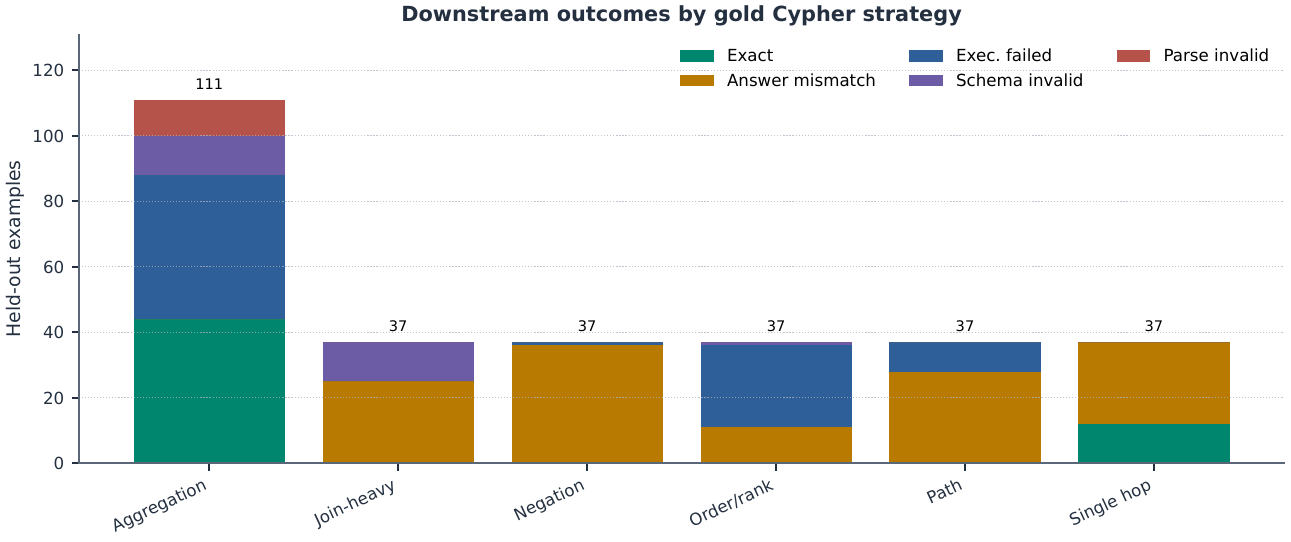}
\caption{Downstream outcomes by gold Cypher strategy on the full held-out test split. The local Qwen3.5-9B baseline fails differently across strategies: aggregation mixes exact matches with execution failures, while join-heavy, negation, path, and ranking examples are dominated by semantically wrong executable queries or execution failures.}
\label{fig:strategy_downstream_errors}
\end{figure*}

\FloatBarrier

\section{Supplementary Evaluator Metrics}

Table~\ref{tab:supported_text_metrics} lists optional text-overlap metrics for debugging answer rendering and near-match behavior. These are not correctness metrics; the reported correctness results use execution accuracy and answer-set F1.

\begin{table}[!htbp]
\centering
\scriptsize
\begin{tabular}{@{}p{0.25\columnwidth}p{0.67\columnwidth}@{}}
\toprule
Metric & Supported use and primary citation \\
\midrule
ROUGE-1/2/L & Reference overlap over serialized answer sets and query strings~\cite{lin-2004-rouge} \\
BLEU & Sentence-level reference overlap with brevity penalty~\cite{papineni-etal-2002-bleu} \\
METEOR & Exact-token precision/recall with fragmentation penalty~\cite{banerjee-lavie-2005-meteor} \\
BERTScore & Optional contextual-embedding similarity~\cite{zhang-etal-2020-bertscore} \\
FrugalScore & Optional efficient learned NLG metric~\cite{kamal-eddine-etal-2022-frugalscore} \\
Cosine & Token-count vector cosine similarity~\cite{manning-etal-2008-ir} \\
Jaro-Winkler & Character-level near-match similarity~\cite{jaro1989record,winkler1990string} \\
Exact match & Raw and normalized string exact match~\cite{rajpurkar-etal-2016-squad} \\
\bottomrule
\end{tabular}
\caption{Supplementary reference-based text metrics supported by the PIPE-Cypher evaluator. These metrics are useful for debugging answer rendering, paraphrase sensitivity, and near-match behavior, but they do not replace execution accuracy or answer-set F1 for Text2Cypher correctness. BERTScore and FrugalScore are optional integrations because they require additional metric/model packages.}
\label{tab:supported_text_metrics}
\end{table}

\FloatBarrier

\section{Governance, Judge Calibration, and Deployment Details}

The deployment details below matter because enterprise failures rarely come from a single model call. They come from missing safety boundaries, unclear value policies, weak audit trails, or no explanation for why an example was accepted. The validator cascade, prompt contracts, automation comparison, and judge audit describe the controls that make PIPE-Cypher a governed local workflow rather than a manual dataset-writing exercise.

\subsection{Validation Cascade}

Table~\ref{tab:validator_cascade} lists the deterministic and judge gates in execution order. It shows how an enterprise deployment keeps unsafe, schema-invalid, empty, or semantically weak examples out of exported benchmarks.

\begin{table}[H]
\centering
\small
\resizebox{\columnwidth}{!}{%
\begin{tabular}{lrr}
\toprule
Gate or ledger & Count & Denominator \\
\midrule
Export accepted & 3,000 & 3,000 \\
Read-only safety & 3,000 & 3,000 \\
Syntax validity & 3,000 & 3,000 \\
Schema/value validity & 3,000 & 3,000 \\
Execution success & 3,000 & 3,000 \\
LLM judge pass & 3,000 & 3,000 \\
Rejected candidates logged & 1,925 & 4,925 \\
\bottomrule
\end{tabular}
}
\caption{PIPE-Cypher validation cascade for the full export and its logged candidate ledger. Unlike Mind the Query, human review is calibration-only.}
\label{tab:validator_cascade}
\end{table}

\subsection{Rewrite and Governance Audits}

Table~\ref{tab:rewrite_audit} addresses rewrite preservation directly. In the reported generation records, generated Cypher was already identical to normalized Cypher, so the accepted benchmark does not rely on a semantics-changing \texttt{RETURN DISTINCT} insertion or projection rewrite. Rewriting is still part of the pipeline, but in this run its role is conservative validation and logging rather than silent semantic alteration.

\begin{table}[H]
\centering
\small
\resizebox{\columnwidth}{!}{%
\begin{tabular}{lr}
\toprule
Rewrite audit property & Value \\
\midrule
Generation records audited & 4,925 \\
Accepted records audited & 3,000 \\
Records changed by normalization & 0 \\
Accepted records changed & 0 \\
RETURN DISTINCT insertions & 0 \\
Accepted RETURN DISTINCT insertions & 0 \\
Rewrite-skip reasons logged & 196 \\
Live comparisons required & 0 \\
Answer-set equality in comparisons & 0 \\
\bottomrule
\end{tabular}
}
\caption{Rewrite prevalence and impact audit over reported generation records. When no generated query differs from its normalized form, no live original/normalized re-execution is required for semantic drift.}
\label{tab:rewrite_audit}
\end{table}

Table~\ref{tab:governance_audit} separates direction, schema/value, syntax/parser, and read-only issues across generation, ablation, and downstream prediction artifacts. The full generation records have no direction failures after validation. Downstream predictions still contain direction errors, which is exactly the kind of graph-specific failure a Cypher benchmark should expose.

\begin{table}[H]
\centering
\small
\resizebox{\columnwidth}{!}{%
\begin{tabular}{lrrrr}
\toprule
Evidence source & Direction & Schema/value & Syntax/parser & Read-only \\
\midrule
Full generation records & 0 & 2 & 0 & 0 \\
Target-size ablations & 260 & 988 & 5 & 0 \\
Downstream predictions & 25 & 9 & 12 & 0 \\
Combined & 285 & 999 & 17 & 0 \\
\bottomrule
\end{tabular}
}
\caption{Governance failure audit. Direction errors, schema/value errors, syntax/parser failures, and read-only violations are counted separately so the appendix shows which Cypher-specific gates do real work.}
\label{tab:governance_audit}
\end{table}

\subsection{Gate-Impact Counterfactual}

Table~\ref{tab:gate_impact} summarizes the first gate that blocks each non-accepted candidate. This gives the counterfactual view missing from a yield-only ablation: it shows what kind of bad example would enter the benchmark if a deployment weakened duplicate/diversity controls, non-empty execution, judge review, schema validation, direction checking, or read-only safety.

\begin{table}[H]
\centering
\small
\resizebox{\columnwidth}{!}{%
\begin{tabular}{lrr}
\toprule
First blocking gate & Candidates & Share \\
\midrule
Accepted & 3,000 & 0.609 \\
Duplicate/diversity & 1,366 & 0.277 \\
Empty result & 486 & 0.099 \\
Judge reject & 67 & 0.014 \\
Schema & 2 & 0.000 \\
Execution failure & 4 & 0.001 \\
\bottomrule
\end{tabular}
}
\caption{Counterfactual first-blocking-gate audit over generation records. The table shows which failure class would enter the benchmark if that gate were removed or weakened.}
\label{tab:gate_impact}
\end{table}

\subsection{Privacy and Redaction Audit}

Table~\ref{tab:redaction_audit} evaluates the redaction policy rather than merely describing it. The audit builds a sensitive-value set from entity bindings, quoted Cypher literals, reverse-grounding literals, and string-valued execution samples. It then applies the configured redactor and exact-matches the raw values against the redacted question, Cypher, entity, and result fields. This does not replace a tenant's PII classifier, but it gives reviewers a measurable privacy check for the value-bearing fields PIPE-Cypher itself creates.

\begin{table}[H]
\centering
\small
\resizebox{\columnwidth}{!}{%
\begin{tabular}{lr}
\toprule
Redaction audit property & Value \\
\midrule
Examples audited & 3,000 \\
Sensitive values checked & 10,956 \\
Examples with sensitive values & 2,970 \\
Examples with residual raw values & 0 \\
Residual raw-value matches & 0 \\
Residual rate per checked value & 0.000 \\
Unique placeholders & 3,754 \\
Reused placeholders & 2,342 \\
Max placeholder frequency & 337 \\
\bottomrule
\end{tabular}
}
\caption{Exact-match redaction audit over value-bearing benchmark surfaces. The audit checks entity bindings, quoted Cypher literals, reverse grounding literals, and string-valued result samples after applying the configured redaction policy.}
\label{tab:redaction_audit}
\end{table}

\subsection{Operational Accounting}

Table~\ref{tab:runtime_accounting} reports local-run accounting from completed generation records. These numbers are for deployment planning: acceptance rate and graph execution latency explain throughput bottlenecks without turning the analysis into a paid-API cost comparison.

\begin{table}[H]
\centering
\small
\resizebox{\columnwidth}{!}{%
\begin{tabular}{lrrrrr}
\toprule
Scope & Records & Accepted & Acceptance & Exec. p50 ms & Exec. p95 ms \\
\midrule
Overall & 4,925 & 3,000 & 0.609 & 11.995 & 32.832 \\
FinBench & 3,405 & 2,000 & 0.587 & 11.837 & 32.398 \\
SNB & 1,520 & 1,000 & 0.658 & 12.420 & 38.558 \\
\bottomrule
\end{tabular}
}
\caption{Operational accounting from completed generation records. These are local-run latency and acceptance diagnostics, not paid-API cost claims.}
\label{tab:runtime_accounting}
\end{table}

\subsection{Prompt Profiles and Contracts}

Table~\ref{tab:prompt_refinement_plan} documents the prompt-profile factors used for ablation planning. The full prompt contracts follow immediately after the table, including the exact LLM-judge system prompt and user prompt template used for reported judge decisions.

\begin{table}[H]
\centering
\small
\resizebox{\columnwidth}{!}{%
\begin{tabular}{lll}
\toprule
Profile & Added constraint & Intended evidence \\
\midrule
Schema-only & Use only visible schema. & Baseline for schema-grounded prompting. \\
Instructions & Exact values, datatype rules, no nested aggregations. & Targets MTQ-style prompt refinements. \\
Examples & Placeholderized retrieved NL-Cypher pairs. & Tests whether examples help without extra governance. \\
Examples + instructions & Few-shot plus explicit rules. & Closest controlled analogue to MTQ Table 5. \\
Full governed & Production-derived Cypher hints, AST-safe rewrites, judge gate. & PIPE-Cypher production setting. \\
\bottomrule
\end{tabular}
}
\caption{Prompt profiles implemented for Mind-the-Query-style prompt-factorial evaluation. Results are reported only for completed, audited target-50-or-larger suites.}
\label{tab:prompt_refinement_plan}
\end{table}

\section{Prompt Contracts}
\label{sec:prompt_contracts}
\label{tab:prompt_contracts}
PIPE-Cypher treats prompts as versioned implementation artifacts. The list summarizes the prompt contracts used for generation, repair, judging, and downstream evaluation; hashes fingerprint the full prompt constants in the codebase.
\begin{enumerate}
\item \textbf{Template generation.} Stage: Workload proposal. SHA-256: \texttt{10b25b}.\par
\textit{Contract.} Schema-only labels, relationships, properties, and categorical values; Realistic enterprise analyst wording; At most two typed slots and JSON-only output
\item \textbf{Reverse binding.} Stage: Graph grounding. SHA-256: \texttt{48e949}.\par
\textit{Contract.} Read-only MATCH/WHERE/RETURN DISTINCT/LIMIT only; Slot variables named exactly as requested; Forward relationship directions from the schema
\item \textbf{Cypher generation.} Stage: Candidate query. SHA-256: \texttt{61c557}.\par
\textit{Contract.} Only schema-visible constructs and observed directions; RETURN DISTINCT for set returns and exact equality for quoted values; Context columns, categorical hints, placeholderized retrieval, and no writes
\item \textbf{Repair.} Stage: Validation feedback. SHA-256: \texttt{56c419}.\par
\textit{Contract.} Preserve question intent while fixing validation or execution issues; Keep query read-only and schema-grounded; Return only corrected Cypher
\item \textbf{LLM judge.} Stage: Quality gate. SHA-256: \texttt{421c7b}.\par
\textit{Contract.} Inputs include question, Cypher, relevant schema excerpt, execution rows, and validation summary; Strict JSON scores for ambiguity, semantic alignment, schema use, and difficulty; Categorical values constrain query literals, not observed result-row values; Pass only useful, unambiguous enterprise benchmark examples
\item \textbf{Downstream Text2Cypher.} Stage: Model evaluation. SHA-256: \texttt{4c07ff}.\par
\textit{Contract.} Read-only Cypher only; Schema-visible constructs and exact direction preservation; RETURN DISTINCT, count/ranking rules, and no explanations
\end{enumerate}
\subsection{LLM Judge Prompt Used in Reported Runs}
The judge runs only after deterministic validation and live execution. For each reviewed example, we fill the template below with the candidate question, Cypher, relevant schema excerpt, sampled execution rows, and validation summary. This is the prompt used for all reported LLM-judge decisions.
\par\smallskip
\begingroup
\scriptsize
\begin{verbatim}
System prompt:
You are an expert benchmark engineer. Return
strict JSON only. No markdown or extra text.

User prompt template:
You are judging whether an NL-to-Cypher
benchmark example is acceptable for an
enterprise benchmark.

Graph schema:
{schema}

Question:
{question}

Cypher:
{cypher}

Execution sample:
{rows}

Validation summary:
{validation}

Return strict JSON with:
- pass: boolean
- ambiguity_score: number from 0 to 1, lower is
better
- semantic_alignment_score: number from 0 to 1
- schema_use_score: number from 0 to 1
- difficulty: one of easy, medium, hard
- failure_reason: short string, empty if pass is
true

Pass only if the question is unambiguous, the
Cypher answers it, the schema use is valid, and
the result would be useful in an enterprise
benchmark.
- Categorical property values in the schema
constrain literal values written in the Cypher
query.
- Do not reject because the execution sample
returns a value that is absent from the
categorical-value list; result rows are observed
graph outputs.
- If deterministic validation says schema use is
valid, lower schema_use_score only when the
Cypher itself uses nonexistent schema elements,
invalid relationship directions, or invalid
literal values.
\end{verbatim}
\endgroup

\subsection{Automation and Calibration}

Table~\ref{tab:effort_automation} gives the deployment contrast with Mind the Query, and Table~\ref{tab:judge_audit_coverage} reports the completed human calibration packet. Human review has not disappeared from the research process. It has moved from a production gate to post-hoc calibration evidence.

\paragraph{Human annotation protocol.}
One external annotator labeled the frozen 80-row audit packet after generation was complete. The packet sampled judge-accepted and judge-rejected candidates across FinBench/SNB categories. For each row, the annotator inspected the NL question, Cypher, graph/category metadata, execution evidence, and judge decision, then filled a binary \texttt{human\_accept} label and optional notes under the rubric in Appendix~\ref{sec:prompt_contracts}: accept only if the question is clear and the Cypher is read-only, schema-grounded, directionally plausible, exact about quoted values, and semantically aligned with the question. The annotator's labels calibrate the local judge and are reported only in aggregate. We do not report protected demographic attributes because a single annotator would be identifiable; recruitment and compensation details are recorded in the Responsible NLP checklist. The annotation protocol was determined exempt by an IRB.

\begin{table}[H]
\centering
\small
\resizebox{\columnwidth}{!}{%
\begin{tabular}{lll}
\toprule
Dimension & Mind the Query & PIPE-Cypher \\
\midrule
Generation review gate & Manual logical review & Deterministic gates + local LLM judge \\
Human effort & Reported 1,400 person-hours & 80-row post-hoc judge calibration audit \\
Private values & Public benchmark values & Configurable sampling and redacted export \\
Refresh & Static dataset release & Rerunnable private benchmark factory \\
Model endpoint & Gemini in their reported pipeline & Local Qwen3.5-9B endpoint \\
\bottomrule
\end{tabular}
}
\caption{Industry deployment contrast with Mind the Query. PIPE-Cypher focuses on private refreshable benchmark generation rather than a one-time public dataset.}
\label{tab:effort_automation}
\end{table}

\begin{table}[H]
\centering
\small
\resizebox{\columnwidth}{!}{%
\begin{tabular}{lr}
\toprule
Audit packet property & Value \\
\midrule
Rows & 80 \\
Human annotators & 1 external annotator \\
IRB status & exempt determination \\
Use of human labels & post-hoc calibration only \\
Judge accept / reject & 40 / 40 \\
FinBench / SNB rows & 48 / 32 \\
Easy / medium rows & 26 / 54 \\
Labeled rows & 80 \\
Calibration status & complete \\
Judge-human agreement / $\kappa$ & 0.800 / 0.600 \\
Judge precision (95\% CI) & 1.000 (0.912--1.000) \\
Judge recall (95\% CI) & 0.714 (0.585--0.816) \\
False-accept rate (95\% CI) & 0.000 (0.000--0.138) \\
False-reject rate (95\% CI) & 0.286 (0.184--0.415) \\
\bottomrule
\end{tabular}
}
\caption{Post-hoc judge calibration packet coverage and judge-human agreement. Human labels calibrate the automated gate after generation and are not used as a generation gate.}
\label{tab:judge_audit_coverage}
\end{table}

\section{Public Artifacts}

The public code repository is available at \url{https://github.com/suraj-ranganath/PIPE-Cypher}. The public-proxy benchmark exports for FinBench/SNB and ICIJ Offshore Leaks are available at \url{https://huggingface.co/datasets/suraj-ranganath/PIPE-Cypher-benchmarks}.

\FloatBarrier
\clearpage
\section{Representative Accepted Examples}
The examples below are selected in stable identifier order from tracked evidence snapshots, one per graph/category cell when available. They show the NL question, accepted Cypher, structural tags, gate status, and a bounded execution-result sample. ICIJ examples are redacted before rendering, preserving query structure while removing value-bearing strings.
\begin{enumerate}
\item \textbf{FinBench / Boolean Existence / easy.}
\textit{Question:} Does account '187743809466009406' have any outgoing transfer?
\begin{quote}\scriptsize\ttfamily\raggedright\sloppy
MATCH (src:Account \{accountId:\\
'187743809466009406'\}) -[:TRANSFER\_TO]-\textgreater{}\\
(:Account)\\
RETURN DISTINCT COUNT(DISTINCT src) \textgreater{} 0\\
AS HasOutgoingTransfer
\end{quote}
\textit{Structure:} single\_hop, aggregation.\par
\textit{Relationships:} TRANSFER\_TO.\par
\textit{Gates:} RO/Syn/Schema/Exec/Judge.\par
\textit{Result sample:} \{HasOutgoingTransfer: True\}; observed rows: 1
\item \textbf{FinBench / Complex Aggregation / medium.}
\textit{Question:} What is the total transferred amount from accounts owned by person 'Gwar'?
\begin{quote}\scriptsize\ttfamily\raggedright\sloppy
MATCH (p:Person \{personName: 'Gwar'\})\\
-[:OWN\_ACCOUNT]-\textgreater{}\\
(src:Account)-[t:TRANSFER\_TO]-\textgreater{}\\
(:Account)\\
RETURN DISTINCT SUM(t.amount) AS\\
TotalTransferredAmount
\end{quote}
\textit{Structure:} join\_heavy, aggregation.\par
\textit{Relationships:} OWN\_ACCOUNT, TRANSFER\_TO.\par
\textit{Gates:} RO/Syn/Schema/Exec/Judge.\par
\textit{Result sample:} \{TotalTransferredAmount: 14972318.41\}; observed rows: 1
\item \textbf{FinBench / Complex Retrieval / easy.}
\textit{Question:} Which accounts received transfers from accounts owned by person 'Zof'?
\begin{quote}\scriptsize\ttfamily\raggedright\sloppy
MATCH (p:Person \{personName: 'Zof'\})\\
-[:OWN\_ACCOUNT]-\textgreater{} (src:Account)\\
-[:TRANSFER\_TO]-\textgreater{} (dst:Account)\\
RETURN DISTINCT dst.accountId AS\\
AccountId, dst.accountType AS\\
AccountType, dst.isBlocked AS IsBlocked\\
LIMIT 300
\end{quote}
\textit{Structure:} join\_heavy, bounded\_result.\par
\textit{Relationships:} OWN\_ACCOUNT, TRANSFER\_TO.\par
\textit{Gates:} RO/Syn/Schema/Exec/Judge.\par
\textit{Result sample:} \{AccountId: 4687402787162554854, AccountType: credit card, IsBlocked: False\}; observed rows: 3
\item \textbf{FinBench / Negation Difference / medium.}
\textit{Question:} Which accounts owned by person 'Kant' have not sent any transfers?
\begin{quote}\scriptsize\ttfamily\raggedright\sloppy
MATCH (p:Person \{personName: 'Kant'\})\\
-[:OWN\_ACCOUNT]-\textgreater{} (a:Account)\\
WHERE NOT (a) -[:TRANSFER\_TO]-\textgreater{}\\
(:Account)\\
RETURN DISTINCT a.accountId AS\\
AccountId, a.accountType AS AccountType,\\
a.isBlocked AS IsBlocked\\
LIMIT 300
\end{quote}
\textit{Structure:} join\_heavy, negation, bounded\_result.\par
\textit{Relationships:} OWN\_ACCOUNT, TRANSFER\_TO.\par
\textit{Gates:} RO/Syn/Schema/Exec/Judge.\par
\textit{Result sample:} \{AccountId: 4739757132830738341, AccountType: merchant account, IsBlocked: False\}; observed rows: 1
\item \textbf{FinBench / Path Temporal / medium.}
\textit{Question:} Which accounts can receive money within two transfer hops from accounts owned by person 'Sossamon'?
\begin{quote}\scriptsize\ttfamily\raggedright\sloppy
MATCH (p:Person \{personName:\\
'Sossamon'\}) -[:OWN\_ACCOUNT]-\textgreater{}\\
(src:Account) -[:TRANSFER\_TO*1..2]-\textgreater{}\\
(dst:Account)\\
RETURN DISTINCT dst.accountId AS\\
AccountId, dst.accountType AS\\
AccountType, dst.isBlocked AS IsBlocked\\
LIMIT 300
\end{quote}
\textit{Structure:} join\_heavy, path, bounded\_result.\par
\textit{Relationships:} OWN\_ACCOUNT, TRANSFER\_TO.\par
\textit{Gates:} RO/Syn/Schema/Exec/Judge.\par
\textit{Result sample:} \{AccountId: 4687402787162554854, AccountType: credit card, IsBlocked: False\}; observed rows: 4
\item \textbf{FinBench / Ranking Topk / medium.}
\textit{Question:} For accounts owned by person 'Barry', which account sent the highest total transfer amount?
\begin{quote}\scriptsize\ttfamily\raggedright\sloppy
MATCH (p:Person \{personName: 'Barry'\})\\
-[:OWN\_ACCOUNT]-\textgreater{}\\
(src:Account)-[t:TRANSFER\_TO]-\textgreater{}\\
(:Account)\\
WITH src, SUM(t.amount) AS totalAmount\\
RETURN DISTINCT src.accountId AS\\
AccountId, src.accountType AS\\
AccountType, src.isBlocked AS IsBlocked,\\
totalAmount\\
ORDER BY totalAmount DESC\\
LIMIT 1
\end{quote}
\textit{Structure:} join\_heavy, aggregation, order\_rank, bounded\_result.\par
\textit{Relationships:} OWN\_ACCOUNT, TRANSFER\_TO.\par
\textit{Gates:} RO/Syn/Schema/Exec/Judge.\par
\textit{Result sample:} \{AccountId: 4732438783436260772, AccountType: internet account, IsBlocked: False\}; observed rows: 1
\item \textbf{FinBench / Simple Aggregation / easy.}
\textit{Question:} How many accounts are owned by person 'Kaewsuktae'?
\begin{quote}\scriptsize\ttfamily\raggedright\sloppy
MATCH (p:Person \{personName:\\
'Kaewsuktae'\}) -[:OWN\_ACCOUNT]-\textgreater{}\\
(a:Account)\\
RETURN DISTINCT COUNT(DISTINCT a) AS\\
AccountCount
\end{quote}
\textit{Structure:} single\_hop, aggregation.\par
\textit{Relationships:} OWN\_ACCOUNT.\par
\textit{Gates:} RO/Syn/Schema/Exec/Judge.\par
\textit{Result sample:} \{AccountCount: 1\}; observed rows: 1
\item \textbf{FinBench / Simple Retrieval / easy.}
\textit{Question:} Which accounts are owned by person 'Barry'?
\begin{quote}\scriptsize\ttfamily\raggedright\sloppy
MATCH (p:Person \{personName: 'Barry'\})\\
-[:OWN\_ACCOUNT]-\textgreater{} (a:Account)\\
RETURN DISTINCT a.accountId AS\\
AccountId, a.accountType AS AccountType,\\
a.isBlocked AS IsBlocked\\
LIMIT 300
\end{quote}
\textit{Structure:} single\_hop, bounded\_result.\par
\textit{Relationships:} OWN\_ACCOUNT.\par
\textit{Gates:} RO/Syn/Schema/Exec/Judge.\par
\textit{Result sample:} \{AccountId: 4732438783436260772, AccountType: internet account, IsBlocked: False\}; observed rows: 1
\item \textbf{ICIJ Offshore Leaks / Boolean Existence / medium.}
\textit{Question:} Does offshore entity 'ENTITY\_VALUE\_1' have a registered address?
\begin{quote}\scriptsize\ttfamily\raggedright\sloppy
MATCH (e:Entity \{name:\\
'ENTITY\_VALUE\_1'\})\\
OPTIONAL MATCH (e)\\
-[:registered\_address]-\textgreater{} (addr:Address)\\
RETURN DISTINCT COUNT(addr) \textgreater{} 0 AS\\
HasRegisteredAddress
\end{quote}
\textit{Structure:} single\_hop, aggregation, optional.\par
\textit{Relationships:} registered\_address.\par
\textit{Gates:} RO/Syn/Schema/Exec/Judge.\par
\textit{Result sample:} \{HasRegisteredAddress: True\}; observed rows: 1
\item \textbf{ICIJ Offshore Leaks / Complex Aggregation / easy.}
\textit{Question:} How many distinct officers are connected to entities in jurisdiction 'JURISDICTION\_VALUE\_1'?
\begin{quote}\scriptsize\ttfamily\raggedright\sloppy
MATCH (o:Officer) -[:officer\_of]-\textgreater{}\\
(e:Entity \{jurisdiction:\\
'JURISDICTION\_VALUE\_1'\})\\
RETURN DISTINCT COUNT(DISTINCT o) AS\\
OfficerCount
\end{quote}
\textit{Structure:} single\_hop, aggregation.\par
\textit{Relationships:} officer\_of.\par
\textit{Gates:} RO/Syn/Schema/Exec/Judge.\par
\textit{Result sample:} \{OfficerCount: 2\}; observed rows: 1
\item \textbf{ICIJ Offshore Leaks / Complex Retrieval / easy.}
\textit{Question:} Which officers share a registered address with offshore entity 'ENTITY\_VALUE\_1'?
\begin{quote}\scriptsize\ttfamily\raggedright\sloppy
MATCH (e:Entity \{name:\\
'ENTITY\_VALUE\_1'\})\\
-[:registered\_address]-\textgreater{} (addr:Address)\\
\textless{}-[:registered\_address]- (o:Officer)\\
RETURN DISTINCT o.node\_id AS OfficerId,\\
o.name AS OfficerName, addr.address AS\\
RegisteredAddress\\
LIMIT 300
\end{quote}
\textit{Structure:} join\_heavy, bounded\_result.\par
\textit{Relationships:} registered\_address.\par
\textit{Gates:} RO/Syn/Schema/Exec/Judge.\par
\textit{Result sample:} \{OfficerId: OFFICER\_ID\_1, OfficerName: OFFICER\_NAME\_1, RegisteredAddress: ADDRESS\_1\}; observed rows: 8
\item \textbf{ICIJ Offshore Leaks / Negation Difference / easy.}
\textit{Question:} Which offshore entities in jurisdiction 'JURISDICTION\_VALUE\_1' do not have a registered address?
\begin{quote}\scriptsize\ttfamily\raggedright\sloppy
MATCH (e:Entity \{jurisdiction:\\
'JURISDICTION\_VALUE\_1'\})\\
WHERE NOT (e) -[:registered\_address]-\textgreater{}\\
(:Address)\\
RETURN DISTINCT e.node\_id AS EntityId,\\
e.name AS EntityName, e.jurisdiction AS\\
Jurisdiction\\
LIMIT 300
\end{quote}
\textit{Structure:} single\_hop, negation, bounded\_result.\par
\textit{Relationships:} registered\_address.\par
\textit{Gates:} RO/Syn/Schema/Exec/Judge.\par
\textit{Result sample:} \{EntityId: ENTITY\_ID\_1, EntityName: ENTITY\_NAME\_1, Jurisdiction: JURISDICTION\_1\}; observed rows: 2
\item \textbf{ICIJ Offshore Leaks / Path Temporal / medium.}
\textit{Question:} Which officers share offshore entities with officer 'OFFICER\_VALUE\_1', and when did each connection start?
\begin{quote}\scriptsize\ttfamily\raggedright\sloppy
MATCH\\
(src:Officer)-[srcRel:officer\_of]-\textgreater{}\\
(entity:Entity) \textless{}-[dstRel:officer\_of]-\\
(dst:Officer)\\
WHERE trim(src.name) = 'OFFICER\_VALUE\_1'\\
AND dst \textless{}\textgreater{} src\\
RETURN DISTINCT dst.node\_id AS\\
OfficerId, dst.name AS OfficerName,\\
entity.name AS SharedEntityName,\\
dstRel.start\_date AS ConnectionStartDate\\
LIMIT 300
\end{quote}
\textit{Structure:} join\_heavy, negation, bounded\_result.\par
\textit{Relationships:} officer\_of.\par
\textit{Gates:} RO/Syn/Schema/Exec/Judge.\par
\textit{Result sample:} \{ConnectionStartDate: DATE\_1, OfficerId: OFFICER\_ID\_1, OfficerName: OFFICER\_NAME\_1\}; observed rows: 15
\item \textbf{ICIJ Offshore Leaks / Ranking Topk / medium.}
\textit{Question:} Which jurisdictions have the most offshore entities?
\begin{quote}\scriptsize\ttfamily\raggedright\sloppy
MATCH (e:Entity)\\
WHERE e.jurisdiction IS NOT NULL\\
WITH e.jurisdiction AS jurisdiction,\\
COUNT(DISTINCT e) AS entityCount\\
RETURN DISTINCT jurisdiction,\\
entityCount\\
ORDER BY entityCount DESC\\
LIMIT 10
\end{quote}
\textit{Structure:} node\_scan, aggregation, order\_rank, negation, bounded\_result.\par
\textit{Relationships:} none.\par
\textit{Gates:} RO/Syn/Schema/Exec/Judge.\par
\textit{Result sample:} \{entityCount: 209634, jurisdiction: JURISDICTION\_1\}; observed rows: 10
\item \textbf{ICIJ Offshore Leaks / Simple Aggregation / easy.}
\textit{Question:} How many offshore entities are connected to officer 'OFFICER\_VALUE\_1'?
\begin{quote}\scriptsize\ttfamily\raggedright\sloppy
MATCH (o:Officer) -[:officer\_of]-\textgreater{}\\
(e:Entity)\\
WHERE trim(o.name) = 'OFFICER\_VALUE\_1'\\
RETURN DISTINCT COUNT(DISTINCT e) AS\\
OffshoreEntityCount
\end{quote}
\textit{Structure:} single\_hop, aggregation.\par
\textit{Relationships:} officer\_of.\par
\textit{Gates:} RO/Syn/Schema/Exec/Judge.\par
\textit{Result sample:} \{OffshoreEntityCount: 1\}; observed rows: 1
\item \textbf{ICIJ Offshore Leaks / Simple Retrieval / easy.}
\textit{Question:} Which offshore entities is officer 'OFFICER\_VALUE\_1' connected to?
\begin{quote}\scriptsize\ttfamily\raggedright\sloppy
MATCH (o:Officer)-[r:officer\_of]-\textgreater{}\\
(e:Entity)\\
WHERE trim(o.name) = 'OFFICER\_VALUE\_1'\\
RETURN DISTINCT e.node\_id AS EntityId,\\
e.name AS EntityName, e.jurisdiction AS\\
Jurisdiction, r.link AS Link\\
LIMIT 300
\end{quote}
\textit{Structure:} single\_hop, bounded\_result.\par
\textit{Relationships:} officer\_of.\par
\textit{Gates:} RO/Syn/Schema/Exec/Judge.\par
\textit{Result sample:} \{EntityId: ENTITY\_ID\_1, EntityName: ENTITY\_NAME\_1, Jurisdiction: JURISDICTION\_1\}; observed rows: 1
\item \textbf{SNB / Boolean Existence / medium.}
\textit{Question:} Does person with id 6597069766828 like any post?
\begin{quote}\scriptsize\ttfamily\raggedright\sloppy
MATCH (p:Person \{id: 6597069766828\})\\
OPTIONAL MATCH (p) -[:LIKES]-\textgreater{}\\
(post:Post)\\
RETURN DISTINCT COUNT(post) \textgreater{} 0 AS\\
LikesAnyPost
\end{quote}
\textit{Structure:} single\_hop, aggregation, optional.\par
\textit{Relationships:} LIKES.\par
\textit{Gates:} RO/Syn/Schema/Exec/Judge.\par
\textit{Result sample:} \{LikesAnyPost: True\}; observed rows: 1
\item \textbf{SNB / Complex Aggregation / medium.}
\textit{Question:} How many distinct posts are in forums joined by person with id 6597069766845?
\begin{quote}\scriptsize\ttfamily\raggedright\sloppy
MATCH (forum:Forum) -[:HAS\_MEMBER]-\textgreater{}\\
(p:Person \{id: 6597069766845\})\\
MATCH (forum) -[:CONTAINER\_OF]-\textgreater{}\\
(post:Post)\\
RETURN DISTINCT COUNT(DISTINCT post) AS\\
JoinedForumPostCount
\end{quote}
\textit{Structure:} join\_heavy, aggregation.\par
\textit{Relationships:} CONTAINER\_OF, HAS\_MEMBER.\par
\textit{Gates:} RO/Syn/Schema/Exec/Judge.\par
\textit{Result sample:} \{JoinedForumPostCount: 1\}; observed rows: 1
\item \textbf{SNB / Complex Retrieval / easy.}
\textit{Question:} Which people are members of forums containing posts tagged 'Manuel\_Noriega'?
\begin{quote}\scriptsize\ttfamily\raggedright\sloppy
MATCH (forum:Forum) -[:HAS\_MEMBER]-\textgreater{}\\
(p:Person), (forum) -[:CONTAINER\_OF]-\textgreater{}\\
(post:Post) -[:HAS\_TAG]-\textgreater{} (tag:Tag\\
\{name: 'Manuel\_Noriega'\})\\
RETURN DISTINCT p.id AS PersonId\\
LIMIT 200
\end{quote}
\textit{Structure:} join\_heavy, bounded\_result.\par
\textit{Relationships:} CONTAINER\_OF, HAS\_MEMBER, HAS\_TAG.\par
\textit{Gates:} RO/Syn/Schema/Exec/Judge.\par
\textit{Result sample:} \{PersonId: 4398046511220\}; observed rows: 19
\item \textbf{SNB / Negation Difference / easy.}
\textit{Question:} Which person records are not linked from any message record through :HAS\_CREATOR?
\begin{quote}\scriptsize\ttfamily\raggedright\sloppy
MATCH (p:Person)\\
WHERE NOT EXISTS((:Message)\\
-[:HAS\_CREATOR]-\textgreater{} (p))\\
RETURN DISTINCT p.id AS PersonId,\\
p.firstName AS PersonFirstName,\\
p.lastName AS PersonLastName
\end{quote}
\textit{Structure:} single\_hop, negation.\par
\textit{Relationships:} HAS\_CREATOR.\par
\textit{Gates:} RO/Syn/Schema/Exec/Judge.\par
\textit{Result sample:} \{PersonFirstName: R., PersonId: 8796093022313, PersonLastName: Rao\}; observed rows: 25
\item \textbf{SNB / Path Temporal / easy.}
\textit{Question:} Which people are within two knows hops of person with id 4398046511136?
\begin{quote}\scriptsize\ttfamily\raggedright\sloppy
MATCH (src:Person \{id: 4398046511136\})\\
-[:KNOWS*1..2]-\textgreater{} (dst:Person)\\
RETURN DISTINCT dst.id AS PersonId,\\
dst.firstName AS FirstName, dst.lastName\\
AS LastName\\
LIMIT 200
\end{quote}
\textit{Structure:} single\_hop, path, bounded\_result.\par
\textit{Relationships:} KNOWS.\par
\textit{Gates:} RO/Syn/Schema/Exec/Judge.\par
\textit{Result sample:} \{FirstName: Rafael, LastName: Fernández, PersonId: 4398046511333\}; observed rows: 25
\item \textbf{SNB / Ranking Topk / medium.}
\textit{Question:} Which city records are linked from the most organisation records through :IS\_LOCATED\_IN?
\begin{quote}\scriptsize\ttfamily\raggedright\sloppy
MATCH (s:Organisation)\\
-[:IS\_LOCATED\_IN]-\textgreater{} (e:City)\\
WITH e, COUNT(DISTINCT s) AS\\
relatedCount\\
RETURN DISTINCT e.id AS TargetId, e.name\\
AS TargetName, relatedCount\\
ORDER BY relatedCount DESC\\
LIMIT 10
\end{quote}
\textit{Structure:} single\_hop, aggregation, order\_rank, bounded\_result.\par
\textit{Relationships:} IS\_LOCATED\_IN.\par
\textit{Gates:} RO/Syn/Schema/Exec/Judge.\par
\textit{Result sample:} \{TargetId: 164, TargetName: Kolkata, relatedCount: 130\}; observed rows: 10
\item \textbf{SNB / Simple Aggregation / easy.}
\textit{Question:} How many posts are tagged 'Vietnam'?
\begin{quote}\scriptsize\ttfamily\raggedright\sloppy
MATCH (post:Post) -[:HAS\_TAG]-\textgreater{} (tag:Tag\\
\{name: 'Vietnam'\})\\
RETURN DISTINCT COUNT(DISTINCT post) AS\\
PostCount
\end{quote}
\textit{Structure:} single\_hop, aggregation.\par
\textit{Relationships:} HAS\_TAG.\par
\textit{Gates:} RO/Syn/Schema/Exec/Judge.\par
\textit{Result sample:} \{PostCount: 1\}; observed rows: 1
\item \textbf{SNB / Simple Retrieval / easy.}
\textit{Question:} Which post IDs did person with id 4398046511124 like?
\begin{quote}\scriptsize\ttfamily\raggedright\sloppy
MATCH (p:Person \{id: 4398046511124\})\\
-[:LIKES]-\textgreater{} (post:Post)\\
RETURN DISTINCT post.id AS PostId\\
LIMIT 200
\end{quote}
\textit{Structure:} single\_hop, bounded\_result.\par
\textit{Relationships:} LIKES.\par
\textit{Gates:} RO/Syn/Schema/Exec/Judge.\par
\textit{Result sample:} \{PostId: 343597385744\}; observed rows: 5
\end{enumerate}

\end{document}